\documentclass{article}
\ifdefined\pdfsuppressptexinfo\pdfsuppressptexinfo=-1\fi 
\usepackage{iclr2027_conference,times}

\usepackage{amsmath,amsfonts,bm}

\def\eqref#1{equation~\ref{#1}}

\def\1{\bm{1}}

\DeclareMathAlphabet{\mathsfit}{\encodingdefault}{\sfdefault}{m}{sl}
\SetMathAlphabet{\mathsfit}{bold}{\encodingdefault}{\sfdefault}{bx}{n}

\usepackage[utf8]{inputenc} 
\usepackage[T1]{fontenc}    
\usepackage{hyperref}       
\usepackage{url}            
\usepackage[table]{xcolor}
\usepackage{colortbl}
\usepackage{booktabs}       
\usepackage{amsfonts}       
\usepackage{nicefrac}       
\usepackage{microtype}      
\usepackage{xcolor}         
\usepackage{amsmath,amssymb,amsthm}
\usepackage{algorithmic}
\usepackage{graphicx}
\usepackage{textcomp}
\usepackage{paralist}
\usepackage{eurosym}
\usepackage{flushend}
\usepackage{soul}
\usepackage{caption}
\usepackage[list=true]{subcaption}
\usepackage{tabu}
\usepackage{wrapfig}
\usepackage{arydshln}
\usepackage[normalem]{ulem}
\usepackage{multirow}
\usepackage{siunitx}
\usepackage{adjustbox}
\usepackage{array}
\usepackage{diagbox}
\usepackage{tikz}
\usetikzlibrary{arrows.meta,positioning,calc,fit,backgrounds,shapes.geometric,shadows}
\usepackage[capitalize]{cleveref}
\usepackage{longtable}

\newtheorem{theorem}{Theorem}
\newtheorem{lemma}{Lemma}
\newtheorem{corollary}{Corollary}
\newtheorem{definition}{Definition}
\newtheorem{proposition}{Proposition}
\theoremstyle{remark}

\theoremstyle{plain}

\providecommand{\todo}[1]{\textit{TODO}}

\title{Steal the Knowledge, Inherit the Mark:\\ TwinMark for Distillation Watermarking\\ via Second Moments and Class Multiplexing}
\hypersetup{hidelinks}

\author{%
  \textbf{Redwanul Karim}$^{1}$ \quad \textbf{Tobias Feigl}$^{1}$ \quad \textbf{Christopher Mutschler}$^{1,2}$ \quad \textbf{Felix Ott}$^{1}$ \\
  $^{1}$Fraunhofer Institute for Integrated Circuits IIS, 90411 Nürnberg, Germany \\
  $^{2}$University of Technology Nürnberg (UTN), 90461 Nürnberg, Germany \\
  \small\texttt{\{redwanul.karim, tobias.feigl, christopher.mutschler, felix.ott\}}\\
  \small\texttt{@iis.fraunhofer.de}
}

\iclrfinalcopy 
\begin{document}

\maketitle
\lhead{Under review as a conference paper at ICLR 2027}

\begin{abstract}
Knowledge distillation can copy a deployed model by training a student on
its logits or features. The student inherits a watermark only through the
output it imitates. Both outputs therefore need a mark, yet most
distillation watermarks cover only one. \textbf{TwinMark} instead
writes one secret payload into what each distillation objective preserves,
the second moment of normalized features and the class-mean logits. Both marks use linear readouts that give sufficient conditions
for bit recovery on fixed audit inputs, while detection is tested
separately under a declared null. In a ten-seed CIFAR-100
study, each mark is detected in every student that imitates its output and
stays at chance otherwise. Marking costs the teacher 1.3 accuracy points
and the audited embedding of feature-distilled students 4.4
nearest-neighbor points. Second-moment bounds certify 50 to 60 of 64
bits in every feature-distilled student, whereas pointwise bounds and the
evaluated logit bounds certify none. Class multiplexing, which signs the
payload per class, raises the logit decoder's rank ceiling and lowers bit
errors at equal energy, and even at $10\times$ over-encoding,
1,024 bits on 100 classes stay detectable in every student. Detection extends to 39 of 40 students of four
other architectures and to segmentation, object
detection and satellite-navigation jammer detection and classification.
These results establish output-matched inheritance under the evaluated
protocols, not resistance to all utility-preserving transformations.
\end{abstract}

\section{Introduction}
\label{label_introduction}

Query access to a deployed model can be enough to copy its behavior
\citep{tramer2016stealing,orekondy2019knockoff}. Knowledge distillation
(KD) does so by training a new \emph{student} network to imitate a
classifier's probabilities \citep{hinton2015distilling} or to match an
embedding service's features \citep{romero2015fitnets,10.1145/3548606.3560586}.
Watermarking lets the owner of the original model, the \emph{teacher},
detect such copies by testing a suspect for a secret signal. Distillation
makes this difficult. The student copies none of the teacher's trained
weights and never sees the owner's key (Section~\ref{sec:preliminaries}),
so any mark it inherits must travel through the outputs it imitates. A
student that imitates the classifier's scores (logits) is never asked to
reproduce a mark in the teacher's features, and one that matches features
is never asked to reproduce a mark in the logits. A model served through
both interfaces therefore needs a mark on each served output, which we
call a \emph{surface}.

Existing watermarks address parts of this problem. Weight watermarks
\citep{uchida2017embedding} and trigger-set watermarks
\citep{adi2018turning,zhang2018protecting} are verified through the
teacher's parameters or its responses to crafted inputs, and a student
distilled on ordinary inputs is not trained to reproduce either.
Watermarks designed for distillation instead place the mark in what the
student learns. Some perturb the served probabilities or logits
\citep{charette2022cosine,zhao2022distillation,lyu2026pdfpufbaseddnnfingerprinting}, others
entangle the mark with task representations
\citep{jia2021entangled,lv2024mea}, and encoder watermarks protect
self-supervised representations
\citep{cong2022sslguard,lv2024sslwm,peng-etal-2023-copying}. Each of them,
however, is read through a single surface. Reading two surfaces is not
new in itself, since DeepSigns \citep{rouhani2019deepsigns} and SEAL
\citep{dai2025sealsubspaceanchoredwatermarksllm} already verify through both hidden layers and outputs.
Formal guarantees, where they exist, answer different questions.
Randomized smoothing covers bounded changes to the parameters
\citep{bansal2022certified}, and the closest analysis relates the single
ownership statistic of the cosine watermark CosWM to the teacher--student
output discrepancy \citep{charette2022cosine}
(Appendix~\ref{app:related_work}). We therefore ask which statistic of
each output can carry one secret multi-bit message, or \emph{payload},
into a distilled student, how many bits it can hold, and which of them the
student's measured imitation error guarantees.

We introduce \textbf{TwinMark}, which writes one $K$-bit payload, derived
from the owner's key, into the statistic that each distillation objective
drives the student to reproduce (Figure~\ref{fig:twinmark-overview}).
Feature-matching KD (FM-KD) trains the student to reproduce the teacher's
normalized features, so the second-moment feature writer (SM-Feat) encodes
each bit in the sign of a key-derived projection of their average outer
product over fixed carrier inputs. Kullback--Leibler KD (KL-KD) instead
matches temperature-softened class probabilities, which determine the
logits up to a common shift. The class-multiplexed logit writer
(Mux-Logit) therefore adds a $P$-dimensional code to the logits in
directions that the probabilities fully determine
(Section~\ref{sec:construction}). It signs the payload differently in each
of the $m$ classes, which we call \emph{class multiplexing}, so the
class-mean logits give $m$ distinct measurements of one payload. This
raises the \emph{rank ceiling}, an upper limit on the bits that a linear
decoder can resolve, from $P<m$ to $mP$. To audit a suspect, the owner
needs only the key and natural inputs, labeled for the logit channel. Each
exposed surface is read as a separate \emph{channel}, and calibrated tests
at a fixed total significance level can detect the payload even when some
bits are wrong (Section~\ref{sec:verification}). Because both decoders are
linear in averages, the measured imitation error also bounds how far each
bit's score can move, so every bit whose teacher score clears that bound
is certified (Section~\ref{sec:theory}).

Our contributions, evaluated in Section~\ref{sec:experiments}, are:
\begin{enumerate}\setlength\itemsep{2pt}
\item \textit{Surface-matched inheritance of one payload.} Each TwinMark
channel is inherited by the students that imitate its surface and not by
the others. Joint training leaves each channel about as recoverable as a
single writer, and training-aware logit marking is inherited more often
than a logit writer added after training.
\item \textit{Certificates that trained students meet.} Because the
feature certificate bounds the second moment after averaging, the
student's deviations on different carriers can cancel. It certifies
detection in every primary FM-KD student, where bounds applied input by
input certify nothing, and a training target chosen from the key alone
removes all feature-bit errors. The logit bounds certify no KL-KD
student, losing every bit once cross-class cancellation is dropped.
\item \textit{Signatures beyond the class count.} Class multiplexing
raises the rank ceiling of the logit decoder. Key-derived signatures up to
ten times the number of classes keep a full-rank decoder and remain
detectable after distillation, and at equal injection energy multiplexing
leaves fewer bits wrong.
\item \textit{Breadth and boundaries.} The mark carries over to
nearly all students of four other architectures and, because SM-Feat
needs only an exposed embedding, to lesion segmentation \citep{8363547},
object detection \citep{Everingham2010} and Global Navigation Satellite
System (GNSS) jammer detection and classification
\citep{heublein_feigl_crpa}. Fixed removal recipes strip the logit mark
only at an accuracy cost that clean models pay too, and a rotation or a
selective change of the public carriers evades the feature verifier.
\end{enumerate}

\section{Problem Setting and Threat Model}
\label{sec:preliminaries}

We now make the setting of Section~\ref{label_introduction} precise. An
owner serves a teacher $T$ through its $m$-class logits, an embedding, or
both. An adversary distills a student $S$ from one of these surfaces, and
the owner audits suspects with a secret key.

\paragraph{Models and objectives.}
For a model $F\in\{T,S\}$ and an input $x$ with label
$y(x)\in\{1,\ldots,m\}$, a comparison head $P_F$ maps the exposed hidden
feature $h_F(x)$ into a $d$-dimensional comparison space declared by
the owner. Write $v_F(x)=P_F(h_F(x))$ and $u_F(x)=v_F(x)/\|v_F(x)\|_2$
for nonzero projection outputs in this common space, $z_F(x)\in\mathbb
R^m$ for the logits and $\pi_F^{(\tau)}=\mathrm{softmax}(z_F/\tau)$ at
temperature $\tau>0$, using the served logits for $T$
(Appendix~\ref{app:notation} lists the notation). KD trains $S$ on
transfer inputs from a distribution $\mathcal D$, possibly with a task
loss, and FM-KD and KL-KD use the canonical objectives
\begin{equation}
\mathcal L_{\mathrm{FD}}=\frac1d\mathbb E_{\mathcal D}\|u_S-u_T\|_2^2,
\qquad
\mathcal L_{\mathrm{KD}}=\tau^2\mathbb E_{\mathcal D}
\mathrm{KL}(\pi_T^{(\tau)}\|\pi_S^{(\tau)}).
\label{eq:kdobj}
\end{equation}
Here $1/d$ averages over feature coordinates, as implemented. Because
probabilities fix the logits only up to a common shift, KL-KD targets the
centered logits (Corollary~\ref{cor:kl-floor},
Appendix~\ref{app:kl-floor-proof}).

\paragraph{Threat model.}
The adversary wants a useful student that cannot be linked to $T$. It
lacks the key and the writers, never queries the verifier, and chooses
the student's architecture, initialization and objective. Key probing,
overwriting, and hard-label or top-$k$ interfaces are out of scope. Each
channel needs only the output its matching objective imitates: full
logits, or full probabilities at a known temperature, for the logit
channel, and the declared embedding for the feature channel. The
inheritance claim for features concerns the declared embedding coordinates and the
responses on fixed, public carriers. A rotated embedding, which the key
alone cannot undo (Proposition~\ref{thm:rot},
Appendix~\ref{app:readout-definitions}), a replaced comparison head and an
input-selective wrapper that alters only the carrier responses are
evasion boundaries of the verifier. Task utility therefore does not
imply that the mark survives, and the guarantees of
Section~\ref{sec:theory} hold only under their output-fidelity conditions.

\section{The TwinMark Construction}
\label{sec:construction}

\begin{figure}[t]
\centering
\includegraphics[width=\textwidth]{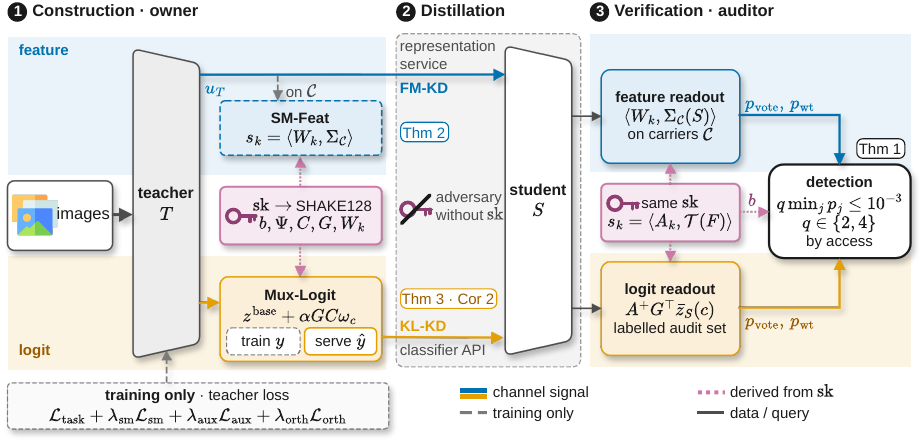}
\caption{Overview. \textbf{(1)} SHAKE128 expands the secret key $\mathrm{sk}$ into the payload $b$ and the readouts. The teacher writes $b$ into its carrier feature second moment (SM-Feat) and logits (Mux-Logit, with $y$ in training and $\hat y$ when served). \textbf{(2)} An adversary without $\mathrm{sk}$ distills a student by FM-KD or KL-KD. \textbf{(3)} The auditor reads each exposed channel with $\mathrm{sk}$ and detects when $q\min_j p_j\le10^{-3}$ ($q{=}2$ p-values for one channel, $4$ for both).}
\label{fig:twinmark-overview}
\end{figure}

This section specifies how the key fixes the payload and how the two
writers of Figure~\ref{fig:twinmark-overview} place it in each surface.

\paragraph{Key derivation.}
SHAKE128, a SHA-3 extendable-output function \citep{NISTFIPS202}, expands the secret key
$\mathrm{sk}$ with one domain tag per object into the payload
$b\in\{\pm1\}^K$, class signatures $\Psi\in\{\pm1\}^{m\times K}$ with
rows $\psi_c$, and a codebook $C\in\mathbb R^{P\times K}$ with unit-norm
columns. The logit frame
$G\in\mathbb R^{m\times P}$ satisfies $G^{\!\top}G=I_P$ and
$G^{\!\top}\mathbf1=0$, with $P\le m-1$ and $\mathbf1$ the all-ones
vector, along which softmax is invariant. The feature
readouts are $W_k=V_kD_kV_k^{\!\top}$, where
$V_k\in\mathbb R^{d\times r}$ has orthonormal columns, $r$ is the
readout rank and $D_k$ is diagonal with entries $\pm1$. The per-class
payload is $\omega_c=b\odot\psi_c$, with $\odot$ the elementwise
product. The fixed carriers $\mathcal C=\{x_\ell\}_{\ell=1}^L$ are not
secret (in classification, the first $L$ training images), and the logit
channel uses a separate labeled audit set (Appendix~\ref{app:training}).

\paragraph{Second-moment feature writer (SM-Feat).}
The feature summary and its bit scores are
\[
\Sigma_C^T=\frac1L\sum_{\ell=1}^Lu_T(x_\ell)u_T(x_\ell)^{\!\top},
\qquad
s_k^{(\mathrm{sm})}(T)=\mathrm{tr}(W_k\Sigma_C^T).
\]
Here $C$ marks the carriers, $\mathrm{tr}$ is the trace, and the same
formulas define $\Sigma_C^F$
and $s_k^{(\mathrm{sm})}(F)$ for any model $F$. The hinge objective
\begin{equation}
\mathcal L_{\mathrm{sm}}=\frac1K\sum_k
\mathrm{ReLU}\bigl(m_{\mathrm{target}}-b_ks_k^{(\mathrm{sm})}(T)\bigr)
\label{eq:Lcov}
\end{equation}
pushes each signed score toward a training target
$m_{\mathrm{target}}$, whereas certificates use the achieved margin
$m_{\mathrm{sm}}$. The key alone also brackets the largest margin that a
trace-one positive semidefinite matrix can give all bits, and a positive
lower bracket fixes a feasible target before training
(Appendix~\ref{sec:target-control}).

\paragraph{Class-multiplexed logit writer (Mux-Logit).}
During training, the teacher adds the label-conditioned payload with
amplitude $\alpha>0$ to the logits $z_T^{\mathrm{base}}$ of its base
head, the ordinary classifier:
\begin{equation}
z_T^{\mathrm{lab}}(x)=z_T^{\mathrm{base}}(x)+\alpha GC\omega_{y(x)}.
\label{eq:logits}
\end{equation}
After projection by $G^{\!\top}$, the class-$c$ mean logit thus carries
$\alpha C\omega_c$. The served teacher replaces $y(x)$ by the base
head's prediction $\hat y(x)=\arg\max_i z_T^{\mathrm{base}}(x)_i$, giving
$z_T^{\mathrm{serve}}=z_T^{\mathrm{base}}+\alpha GC\omega_{\hat y}$
(Definition~\ref{def:serving}, Appendix~\ref{app:readout-definitions}),
which differs from training only on misclassified inputs.

\paragraph{Training objective.}
The teacher minimizes
\begin{equation}
\mathcal L_T=\mathcal L_{\mathrm{task}}
+\lambda_{\mathrm{sm}}\mathcal L_{\mathrm{sm}}
+\lambda_{\mathrm{aux}}\mathcal L_{\mathrm{aux}}
+\lambda_{\mathrm{orth}}\mathcal L_{\mathrm{orth}}
.
\label{eq:LT}
\end{equation}
Here $\mathcal L_{\mathrm{task}}$ is the task loss (cross-entropy of the
label-conditioned logits for classification), the auxiliary loss
$\mathcal L_{\mathrm{aux}}$ fits keyed targets on carrier similarities to
keyed training anchors, and the orthogonality loss
$\mathcal L_{\mathrm{orth}}$ penalizes the leakage
$G^{\!\top}z_T^{\mathrm{base}}$ of base logits into the code subspace
(Appendix~\ref{app:training}).

\section{Verification}
\label{sec:verification}

\paragraph{Decoding.}
The auditor needs only $\mathrm{sk}$, the saved frame $G$ and natural
inputs, with no triggers, learned decoder or fitted threshold. The
feature channel decodes
$\hat b_k=\mathrm{sign}(s_k^{(\mathrm{sm})}(S))$ on the carriers. For the
logit channel, write $\mathbb E_c$ for the empirical mean over labeled
audit inputs of class $c$, with at least one input per class, and $M^+$
for the Moore--Penrose pseudoinverse of a matrix $M$. The auditor
projects each class mean onto the code subspace,
$r_c=G^{\!\top}\mathbb E_c z_S(x)\in\mathbb R^P$, ideally
$\alpha C\omega_c$. For $K<P$, each class determines its own signed copy
$q_c=\alpha^{-1}C^+r_c$, and the decoder undoes the class signs and
averages them into the output score
$s^{(o)}=m^{-1}\sum_c\psi_c\odot q_c$. For $K\ge P$, one class gives only
$P$ numbers for $K$ bits, so the decoder combines all classes as
$s^{(o)}=A^+\mathrm{stack}_c(\alpha^{-1}r_c)$, where
$A=\mathrm{stack}_c(C\,\mathrm{diag}(\psi_c))\in\mathbb R^{mP\times K}$
stacks the $m$ class blocks vertically, which is how class multiplexing
enters the decoder. Both branches decode $\hat b=\mathrm{sign}(s^{(o)})$ with a
fixed tie rule (Definition~\ref{def:verifier},
Appendix~\ref{app:readout-definitions}). In either channel, the bit error
rate $\mathrm{BER}(S,b)$ is the fraction of bits with $\hat b_k\ne b_k$,
and exact recovery means $\mathrm{BER}(S,b)=0$.

\paragraph{Payload tests.}
\label{sec:detection}
Under the null hypothesis $H_0$, the suspect and every choice shaping its
score vector $s$ are independent of a uniformly random payload, as for a
model trained without the key, so each decoded sign matches its bit with
probability one half.
\begin{theorem}[Conditional payload tests and detection]
\label{thm:bertest-cc}
Condition on the suspect, evaluation inputs and labels, non-payload
readout material, decoder, and all choices affecting the score
construction or selection of the tested hypothesis. Under $H_0$,
suppose $b$ remains uniform on $\{\pm1\}^K$ and independent of the
resulting score vector $s$. Use a fixed $+1$ tie rule at zero.
\textbf{(A) Bit agreement.} For
$M=\sum_k\mathbf1\{\operatorname{sign}(s_k)=b_k\}$,
\[
p_{\mathrm{vote}}=\Pr[\mathrm{Binom}(K,1/2)\ge M]
\]
is an exact upper-tail p-value.
\textbf{(B) Weighted agreement.} For $s\ne0$, let
$t=\langle b,s\rangle/\|s\|_2$. The quantity
$p_{\mathrm{wt}}=\exp[-\max(t,0)^2/2]$ is a conservative p-value;
set it to one when $s=0$.
\textbf{(C) Declared test family.} With $q$ tests fixed in advance,
$p_{\mathrm{family}}=\min(1,q\min_jp_j)$ controls Type-I error
at the chosen total level, without independence between tests.
If a recovery certificate guarantees at least $r$ correct bits,
then $p_{\mathrm{vote}}\le\Pr[\mathrm{Binom}(K,1/2)\ge r]$.
This gives a sufficient detection condition even when the
certificate does not cover every bit.
\end{theorem}
The theorem's final clause lets a partial certificate
(Proposition~\ref{prop:partial-recovery}) imply detection. The auditor
fixes its channels before querying and splits the total level
$\alpha_{\mathrm{det}}$ over $q=2$ tests for one surface or $q=4$ for
both, valid even when channels share bits. The theorem is exact when the
payload is independent of all other readout material, which SHAKE
expansion of one finite seed does not guarantee, so we also calibrate the
implemented detector over complete key generation
(Appendix~\ref{app:full-key}) and check it on independently trained
negative models (Section~\ref{sec:experiments}).

\section{From Imitation Error to Certified Bits}
\label{sec:theory}

The tests above hold for any suspect, so we now ask when a distilled
student must make them reject. A \emph{certificate} is a sufficient
condition, evaluated on saved teacher and student outputs at the fixed
audit inputs, that guarantees a decoded bit is correct. It assumes
nothing about the student's architecture or training, but a small KD
loss on the adversary's own inputs does not by itself meet it. It thus
explains an observed recovery, not survival over a prespecified attack
class (proofs in
Appendix~\ref{app:proofs}, procedure in Appendix~\ref{app:design-checklist}).

\paragraph{Certified bits from linear readouts.}
\label{sec:unified}
Both scores are linear in an output statistic, the carrier second moment
or the class-mean logits $\bar z_F(c)=\mathbb E_cz_F$
(Definition~\ref{def:linfunc}, Appendix~\ref{app:readout-definitions}),
so a discrepancy in the statistic bounds each score change. We write
$s_k$ for the $k$th score of either channel.
\begin{proposition}[Certified partial recovery]
\label{prop:partial-recovery}
For either channel, let $\mu_k=b_ks_k(T)$ be the signed teacher
margin and suppose $|s_k(S)-s_k(T)|\le e_k$. Then
\begin{equation}
\mathcal J=\{k:\mu_k>e_k\},\qquad
\mathrm{BER}(S,b)\le1-\frac{|\mathcal J|}{K}.
\label{eq:certified-ber}
\end{equation}
Indeed, every $k\in\mathcal J$ has $b_ks_k(S)>0$.
The bound needs neither independent errors nor a positive margin
on every teacher bit. The following results supply the $e_k$.
\end{proposition}

\paragraph{SM-Feat: \textit{averaging before bounding}.}
\label{sec:covcert-sub}
FM-KD keeps $u_S$ close to $u_T$ on average.
\begin{theorem}[SM-Feat transfer certificate]
\label{thm:covcert}
Let $u_T,u_S$ be unit vectors in a common comparison space and let
$W_k=V_kD_kV_k^{\!\top}$ with $V_k^{\!\top}V_k=I_r$ and diagonal
entries of $D_k$ in $\{\pm1\}$. Suppose the \emph{measured} teacher
margins satisfy $b_ks_k^{(\mathrm{sm})}(T)\ge m_{\mathrm{sm}}>0$.
Writing $d_x=\|u_S(x)-u_T(x)\|_2$,
$\bar\varepsilon_h=\mathbb E_{\mathcal C}d_x$, and
$\Delta\Sigma=\Sigma_C^S-\Sigma_C^T$, for every $k$,
\begin{equation}
\begin{aligned}
|s_k^{(\mathrm{sm})}(S)-s_k^{(\mathrm{sm})}(T)|
&\le e_k^\Sigma:=\|V_k^{\!\top}\Delta\Sigma V_k\|_*\\
&\le \mathbb E_{\mathcal C}[d_x\sqrt{4-d_x^2}]
\le 2\bar\varepsilon_h.
\end{aligned}
\label{eq:cov-sharp}
\end{equation}
Here $\|\cdot\|_*$ is the nuclear norm. Thus
$m_{\mathrm{sm}}>\max_k e_k^\Sigma$ suffices for exact recovery;
$m_{\mathrm{sm}}>2\bar\varepsilon_h$ is a simpler, stronger requirement.
A bit-specific measured margin may replace
$m_{\mathrm{sm}}$ in Eq.~\ref{eq:certified-ber}. Keeping the action
of each $W_k$ before taking norms gives a tighter directional budget
(Proposition~\ref{prop:directional-feature}); bounding the second-moment
change after averaging uniformly refines that budget and retains
cancellation across carriers (Eq.~\ref{eq:second-moment-certificate}).
\end{theorem}
Proposition~\ref{prop:directional-feature} and
Eq.~\ref{eq:second-moment-certificate} appear in
Appendix~\ref{app:covcert-proof}, which also bounds $\bar\varepsilon_h$ by
the FM-KD loss on the carriers. Averaging before bounding lets deviations of different
carriers cancel inside $\Delta\Sigma$, whereas pointwise budgets add
their magnitudes, so $e_k^\Sigma$ can certify bits that pointwise budgets
miss (Section~\ref{sec:experiments}).

\paragraph{Mux-Logit: \textit{residuals and the KL bridge}.}
\label{sec:klcert-cc-sub}
Write $\varepsilon_{\mathrm{top1}}(c)$ for the base head's error rate on
audit inputs of class $c$, $\|\cdot\|_{\mathrm{op}}$ for the operator
norm and $\sigma_{\min}$ for the smallest singular value.
\begin{theorem}[Mux-Logit residual and recovery]
\label{thm:klcert-cc}
Let $\Delta z=z_T^{\mathrm{serve}}-z_S$ and
$\Delta\widetilde z=\Delta z-m^{-1}(\mathbf1^{\!\top}\Delta z)\mathbf1$.
For each class define
\begin{equation}
d_c=\overline{\|\Delta\widetilde z\|}_c:=\mathbb E_c\|\Delta\widetilde z\|_2,
\quad B_c=\|G^{\!\top}\mathbb E_c z_T^{\mathrm{base}}\|_2,
\quad E_c=\alpha\|\mathbb E_c C(\omega_{\hat y}-\omega_c)\|_2.
\label{eq:dz-tilde-bar}
\end{equation}
Then $E_c\le2\alpha\varepsilon_{\mathrm{top1}}(c)\|C\|_{\mathrm{op}}\sqrt K$ and
\begin{equation}
\|r_c-\alpha C\omega_c\|_2\le R_c:=B_c+d_c+E_c.
\label{eq:klcert-cc}
\end{equation}
For $K<P$ and full-column-rank $C$,
$\|s^{(o)}-b\|_\infty\le (m\alpha\sigma_{\min}(C))^{-1}\sum_cR_c$;
a value strictly below one guarantees exact recovery. The stronger
condition $R_c<\alpha\sigma_{\min}(C)$ for every class also recovers
each $\omega_c$ by taking the signs of $q_c$.
For full-column-rank $A$, the stacked decoder instead satisfies
\begin{equation}
\|s^{(o)}-b\|_\infty\le
\frac{\sqrt{\sum_cR_c^2}}{\alpha\sigma_{\min}(A)};
\label{eq:stacked-residual}
\end{equation}
again, a value strictly below one suffices.
\end{theorem}
The residual $R_c$ separates base-logit leakage $B_c$, which
$\mathcal L_{\mathrm{orth}}$ suppresses, the distillation error $d_c$,
the only student-dependent term, and the serving mismatch $E_c$, while
$\sigma_{\min}(C)$ or $\sigma_{\min}(A)$ sets its amplification. A
rowwise budget against the served teacher needs no full rank and is
often tighter (Proposition~\ref{prop:rowwise-logit},
Appendix~\ref{app:klcert-cc-proof}). KL-KD controls probabilities rather
than logits, and Corollary~\ref{cor:kl-floor}
(Appendix~\ref{app:kl-floor-proof}) bridges the two using only the served
teacher. If every teacher probability at the audit temperature $\tau$ is
at least $\delta_c$ on class $c$, then $d_c$ is bounded by an increasing
function of the mean unscaled divergence $\varepsilon_{\mathrm{KL}}(c)$ of
class $c$, approaching
$\tau\sqrt{2\varepsilon_{\mathrm{KL}}(c)/\delta_c}$ as it vanishes, with
no assumption on the student.

\paragraph{Class multiplexing: \textit{rank and conditioning}.}
\label{sec:capacity-cc-sub}
The stacked bound of Theorem~\ref{thm:klcert-cc} divides by
$\sigma_{\min}(A)$, and without class signatures every class would give
the same $P$ measurements, leaving $A$ with rank at most $P$.
\begin{theorem}[Mux-Logit rank and conditional concentration]
\label{thm:capacity-cc}
For any fixed unit-column $C$,
$\mathrm{rank}(A)\le\min(m\,\mathrm{rank}(C),K)\le\min(mP,K)$.
Under independent uniform signs in $\Psi$, independent of $C$, set
$H=C^{\!\top}C-I_K$, $v=\max_k(H^2)_{kk}$, and
$\ell=\log(2K/\beta)$ for $0<\beta<1$. With probability at least
$1-\beta$,
\begin{equation}
\|A^{\!\top}A-mI_K\|_{\mathrm{op}}\le
t_\beta:=\sqrt{2mv\ell}+\tfrac23\|H\|_{\mathrm{op}}\ell.
\label{eq:capacity-bernstein}
\end{equation}
If $t_\beta<m$, then $A$ has full column rank and
$\sigma_{\min}(A)\ge\sqrt{m-t_\beta}$.
\end{theorem}
The rank ceiling thus rises from $P$ to $mP$, which is necessary but not
sufficient. Under independent signatures, an idealization of the SHAKE
expansion, cross-talk between codewords enters $A^{\!\top}A-mI_K$ as
zero-mean terms that cancel on average (proof in
Appendix~\ref{app:capacity-cc-proof}, via matrix Bernstein
\citep{Tropp2012}). As the bound need not be informative for large
payloads, we certify each realized design by its measured rank and
$\sigma_{\min}(A)$, computed from the key alone
(Corollary~\ref{cor:capacity-cond}, Appendix~\ref{app:readout-definitions}).

\section{Experimental Results}
\label{sec:experiments}

\paragraph{Setup.}
The primary study distills CIFAR-100 \citep{Krizhevsky2009LearningML}
ResNet-18 teachers into ResNet-18 students \citep{he2016resnet} over
ten prespecified seeds, with one unscreened owner key per seed.
Teachers train for 60 epochs and students for 30, with shared data,
initialization and schedule. The joint writer uses $K=64$ bits and a
$P=32$ logit code (Appendix~\ref{app:training} lists all settings).
FM-KD trains the student's comparison head, whereas KL-KD, at
$\tau=4.538$, leaves it at initialization. Both add cross-entropy, and neither sees
the key. Each access scenario is tested at total level
$\alpha_{\mathrm{det}}=10^{-3}$. The recipe
was inherited from runs selected on evaluation accuracy, so all
comparisons are fixed-recipe comparisons.

\providecolor{twFeat}{HTML}{0072B2}
\providecolor{twFeatTint}{HTML}{E1EEF7}
\providecolor{twLogit}{HTML}{E69F00}
\providecolor{twLogitTint}{HTML}{FCEFD9}
\begin{table}[t]
\centering\scriptsize
\setlength{\tabcolsep}{2.3pt}
\renewcommand{\arraystretch}{1.08}
\caption{Component study on CIFAR-100 (ResNet-18, $K{=}64$, ten seeds, mean$_{\pm\mathrm{SD}}$ in \%). F/L: feature/logit readout, the unmatched one as a mean. Det.: detections under F/L/dual access at level $10^{-3}$. Shading: the imitated readout. Last block: joint students after removal, from 58.15\% (FM-KD) and 59.12\% (KL-KD). $^\dagger$Fresh key per clean model. $^\ddagger$Native SNR$>$8 rule (teachers 9/10), not a level-$10^{-3}$ test.}
\label{tab:components}
\begin{tabular}{@{}lr|rrrc|rrrc@{}}
\toprule
 & Teacher & \multicolumn{4}{>{\columncolor{twFeatTint}}c|}{\textbf{FM-KD} students (feature matching)} & \multicolumn{4}{>{\columncolor{twLogitTint}}c}{\textbf{KL-KD} students (logit matching)} \\
Method & Acc. & Acc. & F BER & L BER & Det. & Acc. & F BER & L BER & Det. \\
\midrule
Clean & $62.64_{\pm 0.29}$ & $58.19_{\pm 0.27}$ & -- & -- & 0/0/0$^\dagger$ & $59.51_{\pm 0.38}$ & -- & -- & 0/0/0$^\dagger$ \\
Feature only & $61.65_{\pm 0.30}$ & $58.03_{\pm 0.32}$ & \cellcolor{twFeatTint}$2.66_{\pm 3.46}$ & $48.91$ & 10/0/10 & $59.37_{\pm 0.27}$ & $48.12$ & \cellcolor{twLogitTint}$50.00_{\pm 7.37}$ & 0/0/0 \\
Logit only & $61.57_{\pm 0.40}$ & $57.96_{\pm 0.46}$ & \cellcolor{twFeatTint}$50.00_{\pm 6.55}$ & $49.69$ & 0/0/0 & $59.23_{\pm 0.27}$ & $48.91$ & \cellcolor{twLogitTint}$11.41_{\pm 4.04}$ & 0/10/10 \\
Feature + posthoc logit & $61.63_{\pm 0.26}$ & $58.03_{\pm 0.32}$ & \cellcolor{twFeatTint}$2.66_{\pm 3.46}$ & $48.91$ & 10/0/10 & $59.40_{\pm 0.31}$ & $49.06$ & \cellcolor{twLogitTint}$36.41_{\pm 6.12}$ & 0/2/1 \\
Joint, no aux.\ loss & $61.63_{\pm 0.40}$ & $57.79_{\pm 0.30}$ & \cellcolor{twFeatTint}$7.66_{\pm 3.86}$ & $49.38$ & 10/0/10 & $59.02_{\pm 0.35}$ & $49.53$ & \cellcolor{twLogitTint}$9.53_{\pm 4.06}$ & 0/10/10 \\
\textbf{TwinMark} (joint) & $61.35_{\pm 0.41}$ & $58.15_{\pm 0.36}$ & \cellcolor{twFeatTint}$3.44_{\pm 2.74}$ & $49.53$ & 10/0/10 & $59.12_{\pm 0.36}$ & $49.84$ & \cellcolor{twLogitTint}$11.09_{\pm 5.02}$ & 0/10/10 \\
\textbf{TwinMark}, key-only target & $61.19_{\pm 0.36}$ & $58.16_{\pm 0.43}$ & \cellcolor{twFeatTint}$0.00_{\pm 0.00}$ & $49.06$ & 10/0/10 & $58.95_{\pm 0.36}$ & $48.59$ & \cellcolor{twLogitTint}$10.16_{\pm 4.56}$ & 0/10/10 \\
\midrule
CosWM & $62.44_{\pm 0.22}$ & $57.91_{\pm 0.23}$ & \multicolumn{3}{c|}{SNR$>$8: 0/10$^\ddagger$} & $59.38_{\pm 0.29}$ & \multicolumn{3}{c}{SNR$>$8: 0/10$^\ddagger$} \\
\midrule
\multicolumn{10}{@{}l}{\textit{Removal after extraction}} \\
WRT weight shifting & -- & $43.24_{\pm 0.75}$ & \cellcolor{twFeatTint}$3.91_{\pm 2.88}$ & $48.59$ & 10/0/10 & $43.27_{\pm 1.35}$ & $49.06$ & \cellcolor{twLogitTint}$47.66_{\pm 5.76}$ & 0/0/0 \\
Neural Dehydration & -- & $40.07_{\pm 0.93}$ & \cellcolor{twFeatTint}$3.44_{\pm 3.52}$ & $48.75$ & 10/0/10 & $41.31_{\pm 0.66}$ & $49.38$ & \cellcolor{twLogitTint}$22.19_{\pm 5.84}$ & 0/10/9 \\
FTAL & -- & $40.76_{\pm 1.51}$ & \cellcolor{twFeatTint}$3.75_{\pm 2.68}$ & $47.66$ & 10/0/10 & $40.23_{\pm 0.83}$ & $47.81$ & \cellcolor{twLogitTint}$45.16_{\pm 6.77}$ & 0/0/0 \\
RTAL & -- & $44.06_{\pm 1.03}$ & \cellcolor{twFeatTint}$4.84_{\pm 3.25}$ & $49.53$ & 10/0/10 & $43.78_{\pm 1.31}$ & $48.91$ & \cellcolor{twLogitTint}$46.09_{\pm 7.84}$ & 0/0/0 \\
Fine-pruning & -- & $47.07_{\pm 0.77}$ & \cellcolor{twFeatTint}$10.16_{\pm 3.55}$ & $47.81$ & 10/0/10 & $47.40_{\pm 0.95}$ & $47.50$ & \cellcolor{twLogitTint}$45.47_{\pm 6.61}$ & 0/0/0 \\
\bottomrule
\end{tabular}
\end{table}

\begin{table}[t]
\centering
\begin{minipage}[t]{0.485\linewidth}
\centering
\vspace{0pt}
\includegraphics[height=1.8in]{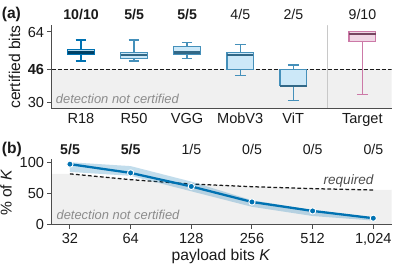}
\captionof{figure}{Second-moment certificates. (a) Certified bits per FM-KD student (boxes: quartiles and range) against the 46 that force vote rejection. Counts: certified detections. Pointwise and logit bounds certify no bit. (b) Certified share of $K$ (median, range) against the share required.}
\label{fig:certificates}
\end{minipage}
\hfill
\begin{minipage}[t]{0.485\linewidth}
\centering
\vspace{0pt}
\captionof{table}{Post-processing attacks on the ten joint CIFAR-100 teachers (\%). Rows test the feature channel, since the logit writer acts at serving time. Det./Cert.: detected/certified seeds at level $10^{-3}$. No clean teacher is flagged. Other datasets and parameters: Appendix~\ref{sec:battery}.}
\label{tab:battery}
\scriptsize
\setlength{\tabcolsep}{2.4pt}
\renewcommand{\arraystretch}{1.06}
\begin{tabular}{@{}lrrcc@{}}
\toprule
Attack on the teacher & Acc. & F BER & Det. & Cert. \\
\midrule
None & $61.36$ & $3.28_{\pm 3.33}$ & 10/10 & 10/10 \\
\multicolumn{5}{@{}l}{\textit{Weights of a stolen teacher}} \\
INT8 quantization & $61.36$ & $3.28_{\pm 3.33}$ & 10/10 & 10/10 \\
INT4 quantization & $51.04$ & $3.91_{\pm 2.78}$ & 10/10 & 10/10 \\
Pruning (50\%) & $57.57$ & $3.44_{\pm 2.83}$ & 10/10 & 10/10 \\
Weight noise & $60.15$ & $3.28_{\pm 2.90}$ & 10/10 & 10/10 \\
Fine-tuning (10 ep.) & $57.23$ & $3.12_{\pm 2.85}$ & 10/10 & 10/10 \\
\multicolumn{5}{@{}l}{\textit{Served embedding (logits unchanged)}} \\
Orthogonal rotation & $61.36$ & $51.41_{\pm 7.81}$ & 0/10 & 0/10 \\
Non-orthogonal warp & $61.36$ & $3.28_{\pm 3.72}$ & 10/10 & 10/10 \\
Feature noise & $61.36$ & $5.47_{\pm 2.47}$ & 10/10 & 10/10 \\
ZCA whitening & $61.36$ & $6.56_{\pm 2.64}$ & 10/10 & 0/10 \\
\bottomrule
\end{tabular}
\end{minipage}
\end{table}

\paragraph{Each objective transfers only the channel it imitates.}
The component study isolates the surface that carries each mark
(Table~\ref{tab:components}). A feature-only mark survives FM-KD in
every seed at 2.66\% BER, but a KL-KD student, whose comparison head
that objective never trains, reads it at 48.12\% BER with no detection.
A logit-only mark shows the converse (11.41\% after
KL-KD, 49.69\% after FM-KD), although FM-KD students fit logits by
cross-entropy. Every unmatched readout of the
joint students also stays at chance, although all students start from
their teacher's initialization, under which distillation can transmit
traits that the imitated outputs do not express \citep{Cloud2026}.

\paragraph{Joint training retains both watermark channels.}
Joint training keeps feature BER at 3.44\% after FM-KD and logit BER
at 11.09\% after KL-KD, detecting all ten seeds, within one
standard deviation (SD) of the single writers. Random class signatures make
the logit code $\psi_c\odot b$ independent of $b$, so an independent logit
payload amounts to another signature draw, and on the first five seeds it
keeps every student detected. Marking costs the teacher 1.29 accuracy points and moves student
classifiers by under half a point against their clean-teacher pairs, but
a fixed cosine 5-nearest-neighbor probe of the verified embedding loses
4.36 points in FM-KD students (Appendix~\ref{sec:component-details}). A
logit writer added after training a feature-only teacher reaches only two
of ten KL-KD students.

\paragraph{A matched distillation watermark is not inherited.}
With its published probability writer and spectral signal-to-noise
ratio (SNR) detector, and our
data, architectures, initializations and budgets, CosWM
\citep{charette2022cosine} flags nine of ten teachers but no student
after either attack (Table~\ref{tab:components}). At a common budget of 100 held-out
queries and conditional level $10^{-3}$, TwinMark detects nine of ten
KL-KD students and CosWM none, even among the six whose teacher it
detects (Appendix Table~\ref{tab:common-query}). CosWM reports an
ownership signal rather than a payload and draws different query classes,
so this compares these implementations and queries, not robustness to
distillation.

\paragraph{Averaging before bounding certifies every primary feature student.}
Evaluated on the saved outputs, the second-moment bound of
Section~\ref{sec:theory} certifies 50 to 60 of 64 bits in each primary
FM-KD student, above the 46 that force the analytic vote test to reject,
so detection is certified in all ten (Figure~\ref{fig:certificates}a),
whereas pointwise norm and directional bounds certify no bit (Appendix
Table~\ref{tab:certificate-coverage}). A hinge target computed from
the key alone before training (Appendix~\ref{sec:target-control})
removes every feature-bit error (Table~\ref{tab:components}), and its
certificates cover all 64 bits in four seeds and detection in nine.

\paragraph{The evaluated logit bounds remain vacuous.}
The rowwise and teacher-floor KL bounds certify no bit, although the
stacked design is well conditioned and every teacher margin is positive.
The observed score changes would certify 569 of 640 bits, but dropping
cross-class cancellation alone certifies none, before any KL step
(Appendix~\ref{sec:certificates-key-only}). KL-KD detection thus rests
on the calibrated tests alone.

\paragraph{Post-processing a stolen teacher leaves the feature mark detectable.}
Nine fixed attacks on the ten CIFAR-100 teachers probe the mark once
the weights leave the owner (Table~\ref{tab:battery}). All five weight
attacks keep detection and certification in every seed, even at a
10-point accuracy cost for four-bit (INT4) quantization. A warp, feature noise
and whitening by zero-phase component analysis (ZCA) of the served
embedding keep detection, although ZCA removes the certificate. An
orthogonal rotation evades the key-only verifier, as Appendix
Proposition~\ref{thm:rot} predicts, and so does a key-independent
replacement of the carrier embeddings of FM-KD students. On CIFAR-10 and
MiniImageNet, only
the rotation and, in two MiniImageNet teachers, ZCA prevent detection
(Appendix~\ref{sec:battery}).

\providecolor{twFeat}{HTML}{0072B2}
\providecolor{twFeatTint}{HTML}{E1EEF7}
\providecolor{twLogit}{HTML}{E69F00}
\providecolor{twLogitTint}{HTML}{FCEFD9}
\begin{table}[t]
\centering\scriptsize
\setlength{\tabcolsep}{2.2pt}
\renewcommand{\arraystretch}{1.08}
\caption{Payload size against class count $m$ and code dimension $P$ (five seeds, mean$_{\pm\mathrm{SD}}$ in \%). CIFAR-100 varies only $K$ ($K{=}64$: first five seeds of Table~\ref{tab:components}). Rank and mean $\sigma_{\min}$: stacked logit decoder, rank $\le\min(K,mP)$. Teacher accuracy: mean. $\Delta$: paired student accuracy change. Det.: matching single access. Shaded: $K>m$.}
\label{tab:payload}
\begin{tabular}{@{}lrrrrr|rrc|rrc@{}}
\toprule
 & & & & & Teacher & \multicolumn{3}{>{\columncolor{twFeatTint}}c|}{\textbf{FM-KD}: feature channel} & \multicolumn{3}{>{\columncolor{twLogitTint}}c}{\textbf{KL-KD}: logit channel} \\
Data ($m$, $P$) & $K$ & $K/m$ & Rank & $\sigma_{\min}$ & Acc. & $\Delta$ & F BER & Det. & $\Delta$ & L BER & Det. \\
\midrule
\multirow{6}{*}{CIFAR-100 (100, 32)} & 32 & 0.32$\times$ & 32 & 9.10 & 61.60 & $-0.23_{\pm 0.66}$ & $1.25_{\pm 1.71}$ & 5/5 & $-0.71_{\pm 0.25}$ & $5.62_{\pm 3.42}$ & 5/5 \\
 & 64 & 0.64$\times$ & 64 & 8.66 & 61.40 & $-0.09_{\pm 0.31}$ & $3.75_{\pm 3.76}$ & 5/5 & $-0.55_{\pm 0.58}$ & $9.69_{\pm 2.04}$ & 5/5 \\
 & 128 & \cellcolor{twLogitTint}1.28$\times$ & 128 & 8.06 & 61.37 & $-0.15_{\pm 0.19}$ & $9.84_{\pm 3.65}$ & 5/5 & $-0.31_{\pm 0.65}$ & $9.22_{\pm 2.02}$ & 5/5 \\
 & 256 & \cellcolor{twLogitTint}2.56$\times$ & 256 & 7.24 & 61.04 & $+0.05_{\pm 0.44}$ & $20.23_{\pm 2.43}$ & 5/5 & $-0.52_{\pm 0.30}$ & $9.84_{\pm 0.97}$ & 5/5 \\
 & 512 & \cellcolor{twLogitTint}5.12$\times$ & 512 & 6.03 & 61.17 & $-0.31_{\pm 0.61}$ & $32.03_{\pm 2.03}$ & 5/5 & $-0.62_{\pm 0.64}$ & $13.67_{\pm 2.09}$ & 5/5 \\
 & 1024 & \cellcolor{twLogitTint}10.24$\times$ & 1024 & 4.36 & 60.82 & $-0.22_{\pm 0.46}$ & $39.34_{\pm 1.15}$ & 5/5 & $-0.63_{\pm 0.45}$ & $16.45_{\pm 1.37}$ & 5/5 \\
\midrule
MiniImageNet (100, 16) & 128 & \cellcolor{twLogitTint}1.28$\times$ & 128 & -- & 59.51 & $-0.37_{\pm 0.84}$ & $17.97_{\pm 4.94}$ & 5/5 & $+0.05_{\pm 0.69}$ & $0.00_{\pm 0.00}$ & 5/5 \\
CIFAR-10 (10, 2) & 32 & \cellcolor{twLogitTint}3.20$\times$ & \textbf{20}\,$<$\,$K$ & -- & 87.19 & $-0.05_{\pm 0.35}$ & $2.50_{\pm 4.07}$ & 5/5 & $-0.29_{\pm 0.26}$ & $31.25_{\pm 5.85}$ & 0/5 \\
\bottomrule
\end{tabular}
\end{table}

\paragraph{Class multiplexing extends the signature beyond the class count.}
At fixed amplitude $\alpha$, the stacked decoder keeps full column rank
up to $K=1{,}024$ while its conditioning degrades (Table~\ref{tab:payload}). Logit BER
after KL-KD rises only to 16.45\%, and every student is detected at every
size, but feature certificates cover detection only up to $K=64$
(Figure~\ref{fig:certificates}b). With the injection energy $\alpha^2K$
held at its primary level, every student is still detected up to
$K=1{,}024$, at 40.5\% logit BER. At that energy and
$K=128$, removing class multiplexing ($\psi_c=\mathbf1$) cuts the rank to
32 and raises logit BER from 17.7\% to 30.2\% in every seed, while both
arms stay detected (Appendix~\ref{sec:capacity-extension}). As detection
power grows with $K$, these results show detectable signatures longer than
the class count, not capacity at a fixed distortion. On CIFAR-10, whose
32-bit payload exceeds $mP=20$, KL-KD gives
$31.25\%$ logit BER and no detections, although rank alone does not
explain this (Appendix~\ref{sec:classification-datasets}).

\providecolor{twFeat}{HTML}{0072B2}
\providecolor{twFeatTint}{HTML}{E1EEF7}
\providecolor{twLogit}{HTML}{E69F00}
\providecolor{twLogitTint}{HTML}{FCEFD9}
\begin{table}[t]
\centering\scriptsize
\setlength{\tabcolsep}{3pt}
\renewcommand{\arraystretch}{1.08}
\caption{Other students and tasks (five seeds, mean$_{\pm\mathrm{SD}}$). CIFAR-100 students distill the first five teacher pairs of Table~\ref{tab:components}. $\Delta$: paired student utility change in points. Det.: matching single access, or dual ($^{\mathrm{D}}$). No clean model is flagged in any task. $^\dagger$Clean-student SD 14.3, no utility claim.}
\label{tab:breadth}
\begin{tabular}{@{}ll|rrc|rrc@{}}
\toprule
 & & \multicolumn{3}{>{\columncolor{twFeatTint}}c|}{\textbf{Feature KD}: feature channel} & \multicolumn{3}{>{\columncolor{twLogitTint}}c}{\textbf{KL-KD}: logit channel} \\
Student model & Data & $\Delta$ & F BER & Det. & $\Delta$ & L BER & Det. \\
\midrule
ResNet-50 & \multirow{4}{*}{CIFAR-100} & $-0.17_{\pm 0.70}$ & $3.44_{\pm 3.56}$ & 5/5 & $-0.19_{\pm 0.66}$ & $5.62_{\pm 2.61}$ & 5/5 \\
VGG-16-BN &  & $+0.11_{\pm 0.30}$ & $3.44_{\pm 3.39}$ & 5/5 & $-0.62_{\pm 0.21}$ & $0.62_{\pm 0.86}$ & 5/5 \\
MobileNetV3-L &  & $0.00_{\pm 0.54}$ & $4.06_{\pm 3.60}$ & 5/5 & $-0.50_{\pm 0.54}$ & $22.81_{\pm 4.07}$ & 5/5 \\
ViT-S/16 (pretrained) &  & $-0.15_{\pm 0.32}$ & $4.06_{\pm 3.60}$ & 5/5 & $-0.67_{\pm 0.22}$ & $27.81_{\pm 7.44}$ & 4/5 \\
\cmidrule(l){1-8}
ViT-S/16 (pretrained) & MiniImageNet & $-0.61_{\pm 0.46}$ & $13.44_{\pm 3.24}$ & 5/5$^{\mathrm{D}}$ & $-0.49_{\pm 0.32}$ & $2.19_{\pm 0.65}$ & 5/5$^{\mathrm{D}}$ \\
\midrule
EffNet-B4 U-Net (Dice) & ISIC 2018 & $-0.04_{\pm 0.18}$ & $3.44_{\pm 3.39}$ & 5/5 & \multicolumn{3}{c}{no logit writer} \\
Swin-T Faster R-CNN (mAP) & VOC 07+12 & $-0.35_{\pm 0.74}$ & $7.81_{\pm 6.54}$ & 5/5 & \multicolumn{3}{c}{no logit writer} \\
ResNet-18, 4-ch.\ (episodic acc.) & GNSS/CRPA & $+15.94_{\pm 11.47}{}^{\dagger}$ & $4.06_{\pm 3.24}$ & 5/5 & \multicolumn{3}{c}{no logit writer} \\
\bottomrule
\end{tabular}
\end{table}

\begin{figure}[t]
\centering
\vspace{-0.2cm}
\includegraphics{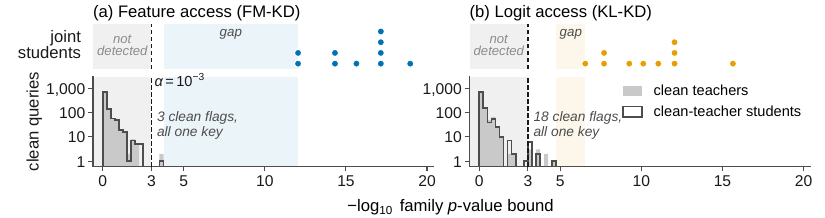}
\vspace{-0.6cm}
\caption{Clean battery queries (histograms) and joint students (dots) under the conditional null.}
\label{fig:null-separation}
\vspace{-0.4cm}
\end{figure}

\paragraph{Inheritance extends to other students and tasks.}
Distilling the first five teacher pairs into four other architectures
gives 20 of 20 feature detections after FM-KD and 19 of 20 logit
detections after KL-KD, with paired accuracy changes from $-0.67$ to
$+0.11$ points (Table~\ref{tab:breadth}), although these students share
no initial weights with their teachers. Logit BER is highest for
MobileNetV3 and the pretrained Vision Transformer (ViT), whose
initialization and batch size also differ, and one ViT seed escapes
detection. MiniImageNet ViT students are all detected, but at
36.5--39.4\% accuracy this is a detection result only. The feature writer also transfers unchanged to
lesion segmentation, object detection and GNSS jammer detection and
classification with a controlled reception pattern antenna (CRPA), where
every student is detected. Segmentation students lose 0.04 Dice points, while
the object-detection students stay too weak in mean average precision
(mAP) to show more than transfer (Appendix~\ref{sec:downstream}).

\paragraph{Fixed removal recipes erase the logit mark while degrading accuracy.}
Five fixed attacks target the extracted students of all ten seeds
(Table~\ref{tab:components}, last block). Weight shifting from the
Watermark Robustness Toolbox (WRT) \citep{lukas2022sok}, fine-tuning all
layers (FTAL), retraining all layers after a classifier reset (RTAL) and
fine-pruning remove every KL-KD logit detection, and basic Neural
Dehydration \citep{10.1145/3658644.3690334} none. WRT costs 14.7--15.9
accuracy points and the others 11 to 19, as much in clean-teacher
students (Appendix Table~\ref{tab:removal-ten}), so these recipes are
not utility-preserving and show vulnerability to fixed attacks, not a
utility cost that removal must pay. The feature mark survives all
five, none of which retrains the verified head, and fine-pruning gives the
only false flag among 100 attacked clean students.

\paragraph{Unattacked clean models are not flagged.}
Ten independently trained clean teachers and their students give no
false detection with one fresh, prespecified key per model, in any
cohort or access scenario. Ten models bound a population
false-positive rate only by 25.89\% (one-sided 95\%), so they check the
executed procedure rather than estimate that rate. A battery of 100 keys per model probes the null
more finely (Figure~\ref{fig:null-separation}). Each dot there is one
joint student, and dots stack where students recover the same number of
payload bits. Every dot lies beyond every clean query, most narrowly
under logit access. The few clean flags come from two keys, so the level
holds on average over keys rather than per key. Rank tests over 1,999
regenerated keys, which need no exact payload independence, reproduce
every analytic decision (Appendix~\ref{sec:negative-controls}).

\section{Conclusion}
\label{label_conclusion}

TwinMark encodes one keyed signature in feature second moments and class-conditioned logits, and under the evaluated distillation protocols each channel is inherited by the students that imitate it. Its feature certificates certify most bits on the fixed audit inputs, while the evaluated logit bounds certify none. Class multiplexing expands the ideal linear measurement rank and lowers bit errors at equal injection energy, but detectable long signatures do not establish information capacity. The feature verifier is coordinate dependent and evadable through its public carriers, and the removal experiments establish no intrinsic utility cost of erasure. These results characterize when watermark evidence follows output imitation and where it stops protecting.




\section*{LLM Usage Disclosure}
Large language models (Claude Opus 5) were used to (i) assist as a general-purpose writing aid for grammar, wording, and clarity, and (ii) draft initial portions of the experiment code, which the authors independently reviewed, tested, and validated. LLMs were not credited as authors and made no autonomous scientific decisions. The authors take full responsibility for the correctness, integrity, and originality of all claims and results.

\bibliographystyle{iclr2027_conference}
\bibliography{main}

\newpage
\appendix

\thispagestyle{plain}
\begingroup
\setlength{\parindent}{0pt}
\newcommand{\apptocsection}[1]{%
  \makebox[0.9cm][l]{\textbf{\ref{#1}}}\textbf{\nameref{#1}}\dotfill\pageref{#1}\par}
\newcommand{\apptocsubsection}[2][]{%
  \hspace*{0.5cm}\makebox[1.0cm][l]{\ref{#2}}\nameref{#2}%
  \if\relax\detokenize{#1}\relax\else\ (#1)\fi\dotfill\pageref{#2}\par}

{\huge\bfseries Appendix}\par
\vspace{1.5em}

{\Large\bfseries Table of Contents}\par
\vspace{0.2em}
\noindent\rule{\linewidth}{0.6pt}
\vspace{0.8em}

\apptocsection{app:related_work}
\apptocsubsection{app:related_work_kd}
\apptocsubsection{app:related_work_watermarking}
\apptocsubsection{app:related_work_watermarking_kd}
\apptocsubsection{app:related_work_formal_guarantees}
\medskip

\apptocsection{app:notation}
\medskip

\apptocsection{app:readout-definitions}
\apptocsubsection{app:decoder-margin}
\medskip

\apptocsection{app:proofs}
\apptocsubsection[Proposition~\ref{thm:rot}]{app:structural}
\apptocsubsection[Theorem~\ref{thm:covcert}]{app:covcert-proof}
\apptocsubsection[Theorem~\ref{thm:klcert-cc}]{app:klcert-cc-proof}
\apptocsubsection[Corollary~\ref{cor:kl-floor}]{app:kl-floor-proof}
\apptocsubsection[Theorem~\ref{thm:bertest-cc} and Proposition~\ref{prop:full-key}]{app:bertest-cc-proof}
\apptocsubsection[Theorem~\ref{thm:capacity-cc} and Corollary~\ref{cor:capacity-cond}]{app:capacity-cc-proof}
\medskip

\apptocsection{app:training}
\apptocsubsection{app:design-checklist}
\medskip

\apptocsection{app:additional_results}
\apptocsubsection{sec:nonvacuity}
\apptocsubsection{sec:coswm-comparison}
\apptocsubsection{sec:certificates-key-only}
\apptocsubsection{sec:battery}
\apptocsubsection{sec:capacity-extension}
\apptocsubsection{sec:crossarch-extension}
\apptocsubsection{sec:wrt-removal}
\apptocsubsection{sec:negative-controls}
\medskip

\apptocsection{sec:downstream}
\medskip

\apptocsection{app:limitations_outlook}
\medskip

\apptocsection{app:licenses_existing_assets}

\vspace{0.8em}
\noindent\rule{\linewidth}{0.6pt}

\endgroup

\newpage

\section{Related Work}
\label{app:related_work}

Watermark survival under distillation depends on two choices: which
teacher outputs the student learns to reproduce, and which statistic
the owner can later inspect. These choices connect the distillation
literature to watermark design and determine the meaning of a
robustness guarantee.

\subsection{Distillation and the Transfer Channel}
\label{app:related_work_kd}

Output-based knowledge distillation trains a student to reproduce a
teacher's softened prediction distribution
\citep{hinton2015distilling}. 
FitNets extends supervision to intermediate representations, using
a learned mapping when teacher and student dimensions differ
\citep{romero2015fitnets}. 
The distinction matters for watermarking: an output-matching loss
directly constrains predictions, whereas feature matching constrains
the selected representation. TwinMark's two writers place the shared
payload in these respective quantities, and its certificates follow
the discrepancy from each quantity to its readout.

Representation transfer also includes attention maps
\citep{zagoruyko2017attention}, 
pairwise distances and triplet angles
\citep{park2019relational}, 
and contrastive objectives
\citep{tian2020contrastive}. 
These methods motivate evaluating more than one notion of student
fidelity. Their training losses do not, by themselves, establish the
second-moment or pointwise discrepancies required by our certificates.
We measure these discrepancies in the declared comparison space,
after any required alignment.

Model extraction makes these transfer mechanisms relevant beyond
cooperative compression. Prediction interfaces can expose enough information
to reproduce model functionality
\citep{tramer2016stealing}, 
including through image--prediction pairs collected without access
to the victim's architecture or training data
\citep{orekondy2019knockoff}. 
Encoders served through embedding interfaces can likewise be stolen by
training a new encoder whose feature vectors match the returned ones
\citep{10.1145/3548606.3560586}.
\citet{jagielski2020high} 
distinguish extraction that achieves high task accuracy from extraction
that closely reproduces the victim's behavior. This distinction also
governs watermark inheritance: accuracy and detector outcomes measure
different properties, while a recovery certificate requires the
specified teacher--student discrepancy on its evaluation inputs.

\subsection{Watermark Observables and Verification Access}
\label{app:related_work_watermarking}

Parameter watermarking embeds a signature through an additional
regularizer on the model weights
\citep{uchida2017embedding}. 
Black-box backdoor watermarking instead makes selected queries produce
owner-chosen responses
\citep{adi2018turning}. 
The first approach verifies an internal parameter statistic, and the
second verifies behavior accessible through a prediction interface.
For distillation, the relevant question is whether the student
preserves the observable used by the detector, even when its
parameters and architecture differ from the teacher's.

DeepSigns already supports both hidden-layer and output-layer
watermarking
\citep{rouhani2019deepsigns}. 
Its hidden-layer procedure embeds multi-bit signatures in activation
distributions, while its output-layer procedure uses selected
key-image/key-label pairs and a statistical response test. This is
close prior work for a framework offering multiple verification
surfaces. TwinMark's distinction is the specific pair of readouts:
signed linear functionals of an uncentered feature second moment,
and class-stratified pseudoinverse decoding of a logit code. Their
analysis tracks how feature and centered-logit errors perturb those
readouts. The comparison therefore concerns the encoded statistics,
decoders, and transfer conditions, rather than the availability of
two interfaces alone.

SEAL also supports latent and output-based verification for large
language models (LLMs)
\citep{dai2025sealsubspaceanchoredwatermarksllm}. It edits factual-anchor representations to
encode bits, reads hidden-state alignment under white-box access,
and uses sentinel-token logits with Bayesian reanchoring under
black-box access. Its evaluated distillation attack trains an
already watermarked student from another teacher, whereas we
study inheritance from a marked teacher by a newly initialized
student. TwinMark's distinction therefore rests on its second-moment
and class-conditioned readouts and their decoder-specific recovery
and detection analysis, not multi-interface verification alone.

Fingerprinting takes a complementary route by identifying an existing
model without first embedding a payload. Conferrable adversarial
examples are designed to transfer from a source model to its stolen
surrogates while avoiding independently trained reference models
\citep{lukas2021conferrable}. 
Such methods emphasize selective behavioral agreement, whereas
TwinMark tests evidence for a preselected secret payload. Both settings
require a comparison against an appropriate null, since agreement
with a source is not, by itself, evidence of a unique training history.

\subsection{Watermarks Designed for Distillation and Encoder Reuse}
\label{app:related_work_watermarking_kd}

Output watermarking can place the signal directly in the responses
used for student training. CosWM adds a cosine perturbation to
teacher probabilities and detects the inherited signal through its
power spectrum, including under ensemble distillation
\citep{charette2022cosine}. 
DRW applies key-dependent probability perturbations to natural language processing models
\citep{zhao2022distillation}. 
PDF adds device-key perturbations to teacher logits and transfers
them through mean-squared-error distillation on raw logits. It recovers device identity
with a learned logit-difference decoder followed by Hamming-distance
matching \citep{lyu2026pdfpufbaseddnnfingerprinting}.
These works establish output perturbation as a mechanism for
distillation-resistant watermarking. TwinMark's logit writer uses a
different encoding and readout: class signatures modulate a shared
finite payload, and a pseudoinverse decodes classwise projected-logit
means. This makes codebook rank and conditioning explicit parts of
the recovery analysis. It also complements the feature writer, which serves suspects that
expose representations rather than classifier outputs.

A second strategy makes watermark behavior share representations with
the task. EWE entangles watermark and task features
\citep{jia2021entangled}. 
MEA-Defender constructs watermark inputs from source-class samples
and constrains their learned behavior to the task domain
\citep{lv2024mea}. 
Margin-based watermarking increases the classification margins of
watermark samples to resist extraction and distillation
\citep{kim2023margin}. 
These methods promote transfer through the training objective.
TwinMark's feature objective instead controls signed second-moment
scores over its carriers. The achieved score margins, rather than
the optimization target alone, determine its certified tolerance
to feature error.

Encoder watermarking further separates representation access from
access to a downstream classifier. SSLGuard maps verification
embeddings through a learned decoder to a secret vector and uses
shadow-encoder training to promote survival under stealing
\citep{cong2022sslguard}. 
SSL-WM instead maps triggered inputs toward a common representation
and verifies the resulting concentration of downstream predicted
labels through an entropy-based test
\citep{lv2024sslwm}. 
SSL-WM thus addresses verification when a downstream head hides the
encoder outputs. EmbMarker protects embedding services of language
models by adding a target embedding to the returned embeddings of texts
that contain trigger words, and verifies a suspect service by comparing
how close its embeddings of trigger and benign texts lie to that target
\citep{peng-etal-2023-copying}. TwinMark's feature channel instead reads a
multi-bit payload from the embeddings of fixed natural carriers, without
trigger inputs. It assumes access to the declared comparison features,
and its logit channel assumes access to the classifier outputs carrying
the class-conditioned code. This access
distinction is necessary for a meaningful comparison across encoder
reuse and classifier distillation.

\subsection{Guarantees and Robustness Evaluation}
\label{app:related_work_formal_guarantees}

Watermark guarantees differ in the quantity they control. Randomized
smoothing certifies watermark robustness within a bounded
$\ell_2$ perturbation of network parameters
\citep{bansal2022certified}. 
CosWM's analysis instead connects its spectral statistic to squared
teacher--student probability discrepancy and ensemble contributions
\citep{charette2022cosine}. 
The latter is a direct precedent for analyzing watermark inheritance
through output fidelity. TwinMark develops this connection for its
two decoders: feature deviation controls a second-moment score,
while classwise centered-logit residuals pass through the
pseudoinverse. The teacher-floor KL bound then supplies a route from
probability divergence to the logit residual without imposing a
student probability floor. These output-level conditions do not
require matching teacher and student parameterizations.

Recovery and statistical evidence answer different questions within
this analysis. A recovery condition determines when decoding preserves
the payload, whereas a calibrated test determines whether the observed
statistic is unusual under a declared null. Neither replaces empirical
evaluation under an explicit attacker and utility constraint.
The systematic evaluation of \citet{lukas2022sok} 
shows how attack selection and tuning can change conclusions about
watermark robustness. MarkErase detects characteristic behavior of
entangled classification watermarks and uses selective distillation
during model extraction to suppress that behavior while retaining the
main task \citep{fei2025markerase}. 
In a different modality, query-based watermark stealing approximately
reverse-engineers an LLM output watermark, enabling both scrubbing and
spoofing \citep{jovanovic2024watermark}. 
Its autoregressive token interface is not a drop-in attack on our
classifier feature/logit interface, but it identifies verifier probing
and secret-mechanism inference as relevant adaptive threats. Neural
Dehydration additionally demonstrates watermark removal by recovering
and unlearning watermark behavior using model internals and limited data
\citep{10.1145/3658644.3690334}. 
These studies motivate reporting the verification interface,
retained student utility, and detection outcome together. TwinMark's
construction and certificates provide the corresponding quantities
for feature and logit transfer, and the following appendices specify their
evaluation conditions and proofs.

\section{Notation and Audit Conditions}
\label{app:notation}

\begin{center}
\small
\begin{tabular}{lp{10cm}}
\toprule
Symbol & Meaning \\
\midrule
$\mathcal C,L$ & Fixed no-augmentation training carriers and their count. Logit class means use the separate labeled audit set. \\
$u_F,d$ & Unit-normalized features in a common $d$-dimensional comparison space. \\
$\Sigma_C^F$ & Empirical uncentered second moment $L^{-1}\sum_{x\in\mathcal C}u_F(x)u_F(x)^{\!\top}$. \\
$\mathbb E_{\mathcal C},\mathbb E_c$ & Empirical means over the carriers and over the audit inputs of class $c$. \\
$W_k,r$ & Symmetric rank-$r$ signed readout, with $\|W_k\|_{\mathrm{op}}=1$. \\
$s_k^{(\mathrm{sm})},s^{(o)}$ & Feature (SM-Feat) bit score and logit (Mux-Logit) score vector. \\
$b,K$ & Secret sign vector and number of encoded coordinates. \\
$m,P$ & Number of classes and logit code dimension, with $P\le m-1$. \\
$C,\Psi,G$ & Unit-column codebook, class-signature matrix, and orthonormal centered-logit frame. \\
$\psi_c,\omega_c$ & Row $c$ of $\Psi$ and the per-class payload $b\odot\psi_c$. \\
$\alpha,\tau$ & Logit writer amplitude and KD temperature, respectively. \\
$\bar\varepsilon_h,d_c$ & Mean feature discrepancy and classwise centered-logit discrepancy on the evaluated inputs. \\
$B_c,E_c,R_c$ & Projected base-logit residual, serving residual, and their bound with distillation error (Theorem~\ref{thm:klcert-cc}). \\
$\delta_c,\varepsilon_{\mathrm{KL}}(c)$ & Minimum served-teacher probability and mean unscaled KL within class $c$. \\
$A,D$ & Stacked writer matrix and regime-selected linear decoder. \\
$M^+$ & Moore--Penrose pseudoinverse of a matrix $M$. \\
$\mathrm{BER}(S,b)$ & Fraction of decoded bits that differ from $b$. \\
$a,\alpha_{\mathrm{det}}$ & An individual test level and the total level allocated within a declared access scenario. \\
\bottomrule
\end{tabular}
\end{center}

\paragraph{Evaluation sets and normalization.}
All discrepancies in a certificate refer to the same inputs as the
corresponding readout. Training-set KD loss does not determine
discrepancy on unqueried carriers. If a class is absent, its mean is
undefined, and the full $m$-class certificate cannot be invoked without
additional samples or a separately declared reduced-class decoder.
This requirement concerns the logit audit set, not the feature carriers.
Teacher and student projection outputs must be
nonzero for the stated unit normalization. For stabilized normalization,
one can use the $2\bar\varepsilon_h$ bound whenever both resulting norms
are at most one, whereas the sharper square-root bound requires unit norms.

\paragraph{Access and scope.}
The logit detector requires class-labeled evaluation inputs and full
logits. Full positive probabilities at a known temperature also
suffice, since $G^{\!\top}z=\tau G^{\!\top}\log p$. Truncation, clipping
or hard labels do not in general preserve this identity. Feature
verification requires access to the declared comparison space.
Discarding that space, adaptive verifier probing, and watermark
overwriting are outside the evaluated attack suite. The deterministic
bounds apply to any student satisfying their measured hypotheses,
and an attack name alone neither establishes nor invalidates a certificate.

\section{Readout Definitions and Decoder Conditioning}
\label{app:readout-definitions}

This appendix states the readout definitions and supporting results
that the main text cites. They use the summaries, serving rule and
decoders of Sections~\ref{sec:construction}--\ref{sec:verification}.

\begin{definition}[Linear-functional readout]
\label{def:linfunc}
Fix a model-output summary $\mathcal T(F)$ and a key-derived linear functional
$A_k$. The continuous bit score is $s_k(F)=\langle A_k,\mathcal T(F)\rangle$.
For SM-Feat, $\mathcal T(F)=\Sigma_C^F$ and $A_k=W_k$ with the Frobenius
inner product. For Mux-Logit, let
$t(F)=\mathrm{stack}_c(\alpha^{-1}G^{\!\top}\bar z_F(c))$.
The score vector is $s^{(o)}(F)=D\,t(F)$, where
\[
D=\begin{cases}
 m^{-1}[\mathrm{diag}(\psi_1)C^+\ \cdots\ \mathrm{diag}(\psi_m)C^+],&K<P,\\
 A^+,&K\ge P,
\end{cases}
\quad A=\mathrm{stack}_c(C\,\mathrm{diag}(\psi_c)).
\]
Thus $A_k$ on the logit side is the $k$th row of $D$.
\end{definition}

\begin{proposition}[Linear-functional unification]
\label{prop:unif}
Both continuous readouts are linear in their respective summaries.
In particular, Definition~\ref{def:linfunc} equals the two deployed
pseudoinverse decoders of Definition~\ref{def:verifier}.
\end{proposition}
This follows by expanding the matrix products. The common form does
not imply independence of the channels or that preserving one leaves
the other unchanged.

\begin{proposition}[Orthogonal equivariance with alignment]
\label{thm:rot}
If features transform by $u(x)\mapsto Ru(x)$, $R\in O(d)$, and the
auditor transforms $W_k\mapsto RW_kR^{\!\top}$, every SM-Feat score
is unchanged. This requires knowledge of $R$, or paired reference
features identifying its action on the span being evaluated. The key
alone does not identify an unknown rotation. An approximate alignment
is covered by the residual deviation in Theorem~\ref{thm:covcert}.
\end{proposition}

\begin{definition}[Mux-Logit writer]
\label{def:cc-writer}
Derive $b\in\{\pm1\}^K$, $\Psi\in\{\pm1\}^{m\times K}$,
unit-column $C\in\mathbb R^{P\times K}$, and
$G\in\mathbb R^{m\times P}$ with $G^{\!\top}G=I_P$,
$G^{\!\top}\mathbf1=0$ ($P\le m-1$). With
$\omega_c=b\odot\psi_c$ and amplitude $\alpha>0$, the label-conditioned training writer is
$z_T^{\mathrm{lab}}(x)=z_T^{\mathrm{base}}(x)+\alpha GC\omega_{y(x)}$.
\end{definition}

\begin{definition}[Argmax-conditioned serving]
\label{def:serving}
The served teacher is
$z_T^{\mathrm{serve}}(x)=z_T^{\mathrm{base}}(x)+\alpha GC\omega_{\hat y(x)}$,
where $\hat y(x)=\arg\max_i z_T^{\mathrm{base}}(x)_i$ with a fixed
tie rule. Define $\varepsilon_{\mathrm{top1}}(c)$ as the base head's
error within evaluation class $c$. Training and served logits agree
on every correctly classified input, and may also agree on others.
This does not assert that the watermarked argmax equals the base argmax.
\end{definition}

\begin{definition}[Class-stratified verifier, two decoders]
\label{def:verifier}
On a labeled evaluation set with at least one sample in each class,
let $r_c=G^{\!\top}\mathbb E_c z_S(x)\in\mathbb R^P$.
For $K<P$, compute $q_c=\alpha^{-1}C^+r_c$,
$s^{(o)}=m^{-1}\sum_c\psi_c\odot q_c$, and
$\hat b=\mathrm{sign}(s^{(o)})$.
For $K\ge P$, form $y=\mathrm{stack}_c(\alpha^{-1}r_c)$ and
$s^{(o)}=A^+y$, then take its signs. Zero scores use a fixed tie rule.
Both decoders are defined even for deficient-rank matrices. The
full-message bounds use full column rank; the per-bit teacher-reference
bound does not. Expectations here and in the certificates are empirical
averages on this same set; population versions require the corresponding
integrability assumptions.
\end{definition}

\subsection{Decoder Margin of the Stacked Design}
\label{app:decoder-margin}

Theorem~\ref{thm:capacity-cc} (Section~\ref{sec:capacity-cc-sub}) bounds
the rank and conditioning of the stacked design $A$. The following
corollary converts a measured smallest singular value into a decoder
margin.

\begin{corollary}[Conditioned decoder margin]
\label{cor:capacity-cond}
For full-column-rank $A$ and $y=Ab+\eta$, the continuous decoder
$\widetilde b=A^+y$ satisfies
\begin{equation}
\|\widetilde b-b\|_\infty\le
\min\left\{\frac{\|\eta\|_2}{\sigma_{\min}(A)},\;
\|A^+\|_{\infty\to\infty}\|\eta\|_\infty\right\},
\label{eq:decoder-margin}
\end{equation}
where $\|A^+\|_{\infty\to\infty}$ is the maximum absolute row sum.
Either bound strictly below one guarantees
$\mathrm{sign}(\widetilde b)=b$. A spectral-only infinity-noise
condition is $\sqrt{mP}\|\eta\|_\infty<\sigma_{\min}(A)$.
The strict threshold ensures that no decoded coordinate is zero.
\end{corollary}

\section{Proofs}
\label{app:proofs}

\subsection{Orthogonal Equivariance}
\label{app:structural}

\begin{proof}[Proof of Proposition~\ref{thm:rot}]
Under $u\mapsto Ru$, the second moment becomes $R\Sigma R^{\!\top}$.
Rotating the readout as well gives
\[
\mathrm{tr}\bigl((RW_kR^{\!\top})(R\Sigma R^{\!\top})\bigr)
=\mathrm{tr}(RW_k\Sigma R^{\!\top})
=\mathrm{tr}(W_k\Sigma).
\]
Paired reference vectors spanning the evaluated feature span determine
the action of an exact orthogonal map on that span. A Procrustes fit
can recover that action when the pairs are noiseless and related by
the same map; its action on an unobserved orthogonal complement need
not be identifiable. For an approximate fit, apply the feature
certificate to the aligned vectors.
\end{proof}

\subsection{Feature Recovery and Its Relation to Feature Matching}
\label{app:covcert-proof}

\begin{proof}[Proof of Theorem~\ref{thm:covcert}]
Trace duality with $\|D_k\|_{\mathrm{op}}=1$ gives
\[
|\mathrm{tr}(D_kV_k^{\!\top}\Delta\Sigma V_k)|
\le\|V_k^{\!\top}\Delta\Sigma V_k\|_*.
\]
Compression by $V_k$ cannot increase the nuclear norm. Convexity
therefore bounds the right-hand side by
$\mathbb E_{\mathcal C}\|u_Su_S^{\!\top}-u_Tu_T^{\!\top}\|_*$.
For unit vectors this rank-two matrix has eigenvalues
$\pm\sqrt{1-(u_S^{\!\top}u_T)^2}$, with zero values omitted.
Its nuclear norm is $d_x\sqrt{4-d_x^2}\le2d_x$, proving the
entire chain in Eq.~\ref{eq:cov-sharp}. For every bit,
\[
b_ks_k^{(\mathrm{sm})}(S)
\ge b_ks_k^{(\mathrm{sm})}(T)
-|s_k^{(\mathrm{sm})}(S)-s_k^{(\mathrm{sm})}(T)|>0
\]
under the stated margin condition, proving sign recovery.
Finally $\|W_k\|_{\mathrm{op}}=1$ gives
$|u^{\!\top}W_ku|\le1$, which bounds every averaged readout.
\end{proof}

\paragraph{Relating normalized and raw feature matching.}
The implemented normalized-feature loss of Eq.~\ref{eq:kdobj}, evaluated
on the carriers as $\mathcal L_{\mathrm{FD},\mathcal C}$, satisfies
$\bar\varepsilon_h\le\sqrt{d\mathcal L_{\mathrm{FD},\mathcal C}}$
by Cauchy--Schwarz. For raw-feature matching, write
$r_F=\|v_F\|_2>0$ and
$\mathcal L_{\mathrm{raw},\mathcal C}=\mathbb E_\mathcal C\|v_S-v_T\|_2^2$.
The radial--angular identity
\[
\|v_S-v_T\|_2^2=(r_S-r_T)^2+r_Sr_T\|u_S-u_T\|_2^2
\]
gives $\|u_S-u_T\|_2\le\|v_S-v_T\|_2/\sqrt{r_Sr_T}$.
If both raw norms are at least $\gamma>0$, averaging gives
$\bar\varepsilon_h\le\sqrt{\mathcal L_{\mathrm{raw},\mathcal C}}/\gamma$.
Equal raw norms attain the pointwise constant; direct carrier
measurements can further tighten either RMS-based bound.

\begin{proposition}[Directional feature certificate]
\label{prop:directional-feature}
Under the feature readout assumptions of Theorem~\ref{thm:covcert},
write $\delta_x=u_S(x)-u_T(x)$ and $q_x=u_S(x)+u_T(x)$. Set
\begin{equation}
e_k^F=\mathbb E_{\mathcal C}\min\{
 \|\delta_x\|_2\|W_kq_x\|_2,
 \|W_k\delta_x\|_2\|q_x\|_2\}.
\label{eq:directional-feature}
\end{equation}
Then $|s_k^{(\mathrm{sm})}(S)-s_k^{(\mathrm{sm})}(T)|\le e_k^F$
and
\[
e_k^F\le\mathbb E_{\mathcal C}
 [d_x\sqrt{4-d_x^2}]\le2\bar\varepsilon_h.
\]
Thus Eq.~\ref{eq:certified-ber} holds with
$\mu_k=b_ks_k^{(\mathrm{sm})}(T)$ and $e_k=e_k^F$.
\end{proposition}

\begin{proof}
Symmetry gives the exact identity
$u_S^{\!\top}W_ku_S-u_T^{\!\top}W_ku_T
=\delta_x^{\!\top}W_kq_x$.
Applying Cauchy--Schwarz with $W_k$ on either vector bounds the
absolute value by the minimum in Eq.~\ref{eq:directional-feature}.
Average and use the triangle inequality. Since
$\|W_k\|_{\mathrm{op}}\le1$ and the two features are unit vectors,
both products are bounded by $d_x\sqrt{4-d_x^2}$.
Subtracting this error bound from each signed teacher score and
counting the positive lower bounds proves the BER statement.
\end{proof}

\paragraph{Averaging before bounding.}
The directional bound averages absolute deviations and can lose
cancellation across carriers. An alternative uses
$\Delta\Sigma=\Sigma_C^S-\Sigma_C^T$ directly:
\begin{equation}
|s_k^{(\mathrm{sm})}(S)-s_k^{(\mathrm{sm})}(T)|
\le e_k^{\Sigma}:=\|V_k^{\!\top}\Delta\Sigma V_k\|_*.
\label{eq:second-moment-certificate}
\end{equation}
Here $\|\cdot\|_*$ is the nuclear norm. Indeed, the score change
is $\mathrm{tr}(D_kV_k^{\!\top}\Delta\Sigma V_k)$, and trace
duality with $\|D_k\|_{\mathrm{op}}=1$ proves the inequality.
Moreover, $e_k^{\Sigma}\le e_k^F$: averaging before bounding
uniformly refines the directional certificate for these readouts.
To see this, any vectors $a,b$ satisfy
\[
\|aa^{\!\top}-bb^{\!\top}\|_*
=\|a-b\|_2\|a+b\|_2.
\]
Writing $h=a-b$ and $q=a+b$, the matrix is
$(hq^{\!\top}+qh^{\!\top})/2$, whose at most two nonzero
eigenvalues are $(h^{\!\top}q\pm\|h\|_2\|q\|_2)/2$,
with zero values omitted.
Cauchy--Schwarz gives the stated nuclear norm; the identity
also holds when the vectors are dependent or either factor is zero.
Apply this identity to $a=V_k^{\!\top}u_S$ and
$b=V_k^{\!\top}u_T$, then use convexity of the nuclear norm:
\[
e_k^{\Sigma}
\le\mathbb E_{\mathcal C}
 [\|V_k^{\!\top}\delta_x\|_2\|V_k^{\!\top}q_x\|_2]
\le e_k^F.
\]
The last step uses $\|V_k^{\!\top}v\|_2\le\|v\|_2$
and $\|W_kv\|_2=\|V_k^{\!\top}v\|_2$.
Thus $e_k^{\Sigma}$ can directly replace $e_k^F$ in
Eq.~\ref{eq:certified-ber}.

For a general numerically stored symmetric readout $W_k$, the bound
$\||W_k|^{1/2}\Delta\Sigma|W_k|^{1/2}\|_*$ retains all of its
eigenmodes rather than assuming exact orthogonality after rounding.
It follows from trace duality with the spectral sign of $W_k$.
This generalized bound need not be smaller than the directional bound
when the nonzero eigenvalue magnitudes differ, so their minimum
remains valid for the finite-precision implementation.
This is a second-moment transfer condition, not a consequence of
small training loss on different inputs.

\subsection{Logit Residuals and Decoder Recovery}
\label{app:klcert-cc-proof}

\begin{proof}[Proof of Theorem~\ref{thm:klcert-cc}]
Condition on evaluation class $c$. The served writer satisfies
\[
G^{\!\top}\mathbb E_c z_T^{\mathrm{serve}}
=G^{\!\top}\mathbb E_c z_T^{\mathrm{base}}+\alpha C\omega_c
+\underbrace{\alpha\mathbb E_c C(\omega_{\hat y}-\omega_c)}_{\Delta_{\mathrm{serve}}(c)\in\mathbb R^P}.
\]
Since $G^{\!\top}\mathbf1=0$ and $\|G^{\!\top}\|_{\mathrm{op}}=1$,
\[
\|G^{\!\top}\mathbb E_c\Delta z\|_2
=\|G^{\!\top}\mathbb E_c\Delta\widetilde z\|_2
\le\mathbb E_c\|\Delta\widetilde z\|_2=d_c.
\]
Subtract the teacher--student difference and use the triangle
inequality to obtain $R_c=B_c+d_c+E_c$. The serving difference
vanishes on correctly classified base-head inputs, and
$\|\omega_{\hat y}-\omega_c\|_2\le2\sqrt K$, proving the bound on $E_c$.

Let $e_c=r_c-\alpha C\omega_c$. For full-column-rank $C$,
\[
q_c=\omega_c+\alpha^{-1}C^+e_c,\qquad
s^{(o)}-b=\frac1{m\alpha}\sum_c\mathrm{diag}(\psi_c)C^+e_c.
\]
Each diagonal sign matrix preserves the Euclidean norm. Thus
$\|s^{(o)}-b\|_\infty\le\|s^{(o)}-b\|_2
\le\sum_cR_c/(m\alpha\sigma_{\min}(C))$.

For the stacked branch, $y=Ab+\eta$ with
$\eta=\mathrm{stack}_c(e_c/\alpha)$, so
\[
\|\eta\|_2^2=\alpha^{-2}\sum_c\|e_c\|_2^2
\le\alpha^{-2}\sum_cR_c^2.
\]
When $A^+A=I_K$, $s^{(o)}-b=A^+\eta$, yielding
Eq.~\ref{eq:stacked-residual}. If any continuous coordinate error has
magnitude strictly below one, it cannot change the sign of a target
coordinate in $\{\pm1\}$.
\end{proof}

\begin{proposition}[Rowwise logit transfer and writer certificates]
\label{prop:rowwise-logit}
Fix the declared evaluation classes and either linear decoder $D$
from Definition~\ref{def:linfunc}. Write $D_{k,c}\in\mathbb R^P$
for the class-$c$ block of its $k$th row and define
\[
\rho_c=\|G^{\!\top}(\bar z_S(c)-\bar z_T^{\mathrm{serve}}(c))\|_2
\le d_c,\qquad L_{k,c}=\|D_{k,c}\|_2.
\]
The teacher-reference certificate is
\begin{equation}
b_ks_k^{(o)}(S)\ge
\underbrace{b_ks_k^{(o)}(T^{\mathrm{serve}})}_{\mu_k^T}
-\underbrace{\alpha^{-1}\sum_cL_{k,c}\rho_c}_{e_k^T}.
\label{eq:teacher-rowwise}
\end{equation}
The writer-reference certificate is
\begin{equation}
b_ks_k^{(o)}(S)\ge
\underbrace{b_k(DAb)_k}_{\mu_k^A}
-\underbrace{\alpha^{-1}\sum_cL_{k,c}(B_c+E_c+\rho_c)}_{e_k^A}.
\label{eq:writer-rowwise}
\end{equation}
Each bound gives Eq.~\ref{eq:certified-ber}. Their maximum may also
be used coordinatewise. Both statements hold for a fixed
deficient-rank design: in that case $\mu_k^A$ must be evaluated rather
than replaced by one. If $DA=I_K$, every $\mu_k^A=1$.
\end{proposition}

\begin{proof}
The difference between the two decoded score vectors equals
\[
s^{(o)}(S)-s^{(o)}(T^{\mathrm{serve}})
=\alpha^{-1}D\,\mathrm{stack}_c
 [G^{\!\top}(\bar z_S(c)-\bar z_T^{\mathrm{serve}}(c))].
\]
Blockwise Cauchy--Schwarz proves Eq.~\ref{eq:teacher-rowwise}.
For Eq.~\ref{eq:writer-rowwise}, use the exact decomposition
\[
G^{\!\top}\bar z_S(c)-\alpha C\omega_c
=G^{\!\top}\bar z_T^{\mathrm{base}}(c)
+\Delta_{\mathrm{serve}}(c)
+G^{\!\top}(\bar z_S(c)-\bar z_T^{\mathrm{serve}}(c)).
\]
The norm is at most $B_c+E_c+\rho_c$. Hence
$s^{(o)}(S)=DAb+D\eta$, where each block of $\eta$ has norm
at most $(B_c+E_c+\rho_c)/\alpha$. A second blockwise
Cauchy--Schwarz inequality proves the result. Neither argument uses
full rank. For a full-column-rank stacked design,
\[
\sum_cL_{k,c}R_c
\le\|D_{k,:}\|_2\sqrt{\sum_cR_c^2}
\le\frac{\sqrt{\sum_cR_c^2}}{\sigma_{\min}(A)}.
\]
Thus the rowwise writer error is no larger than the original spectral
error bound. For the per-class branch,
$L_{k,c}=\|(C^+)_{k,:}\|_2/m$, which likewise improves or matches
the spectral bound using $\sigma_{\min}(C)^{-1}$.
\end{proof}

The teacher-reference form measures the margin of the actual served
teacher, so its base-head leakage and serving mismatch are already
included in $\mu_k^T$. The writer-reference form separates these
contributions and exposes the effect of codebook conditioning. They
support complementary audits of the same decoder, without altering
either writer.

\subsection{A KL Bound Using Only a Teacher Probability Floor}
\label{app:kl-floor-proof}

Section~\ref{sec:theory} uses the following bridge from the classwise
divergence that KL-KD minimizes to the centered-logit error $d_c$ of
Theorem~\ref{thm:klcert-cc}.

\begin{corollary}[KL-to-centered-logit bound with a teacher floor]
\label{cor:kl-floor}
At a fixed $\tau>0$, let $p_T=\mathrm{softmax}(z_T^{\mathrm{serve}}/\tau)$,
$p_S=\mathrm{softmax}(z_S/\tau)$ and suppose
$\min_i p_T(x)_i\ge\delta_c>0$ for every evaluated input of class $c$.
For the \emph{unscaled} divergence
$\varepsilon_{\mathrm{KL}}(c)=\mathbb E_c\mathrm{KL}(p_T\|p_S)$,
\begin{equation}
d_c\le J_c:=\frac{\tau}{\sqrt2}
h^{-1}\!\left(\frac{2\varepsilon_{\mathrm{KL}}(c)}{\delta_c}\right),
\qquad h(t)=t-1+e^{-t},\quad t\ge0.
\label{eq:kl-floor}
\end{equation}
Thus $d_c$ in Theorem~\ref{thm:klcert-cc} can be replaced by $J_c$,
or by $\min(d_c,J_c)$ if both are measured. No student probability
floor or logit-gap assumption is needed. If the reported loss includes
$\tau^2$, divide that loss by $\tau^2$ before using it here.
\end{corollary}
The bound approaches $\tau\sqrt{2\varepsilon_{\mathrm{KL}}(c)/\delta_c}$
as the divergence tends to zero. The floor must hold for the served
teacher at the actual audit temperature. A small training-set average
KL does not establish classwise budgets on a different audit set.
Using each input's own floor before averaging sharpens this bound
(Corollary~\ref{cor:pointwise-kl-floor}).

\begin{lemma}[Fisher lower bound on centered vectors]
\label{lem:mintoeig}
For a probability vector $p$ with $\min_i p_i\ge\delta$ and
$v\perp\mathbf1$,
$v^{\!\top}(\mathrm{diag}(p)-pp^{\!\top})v\ge\delta\|v\|_2^2$.
\end{lemma}
\begin{proof}
The variance identity gives
$v^{\!\top}F_pv=\min_a\sum_i p_i(v_i-a)^2
\ge\delta\min_a\sum_i(v_i-a)^2=\delta\|v\|_2^2$,
where the last equality uses the zero coordinate mean.
\end{proof}

\begin{proof}[Proof of Corollary~\ref{cor:kl-floor}]
Fix an input, let $p=p_T$, and set
$v=\Delta\widetilde z/\tau$, $w=z_T^{\mathrm{serve}}/\tau$.
Centering changes logits by a scalar, so
$p_S=\mathrm{softmax}(w-v)$. For
$f(w)=\log\sum_i e^{w_i}$, Taylor's integral identity gives
\begin{equation}
\mathrm{KL}(p\|p_S)
=\int_0^1(1-t)\,v^{\!\top}F_{p_t}v\,dt,
\qquad p_t=\mathrm{softmax}(w-tv).
\label{eq:kl-integral}
\end{equation}
Writing $R=\max_i v_i-\min_i v_i$, we have
\[
(p_t)_i=\frac{p_i e^{-tv_i}}{\sum_jp_j e^{-tv_j}}
\ge p_i e^{-tR}\ge\delta_c e^{-tR}.
\]
Moreover $R\le\sqrt2\|v\|_2=:s$. Applying
Lemma~\ref{lem:mintoeig} inside the integral yields, for $s>0$,
\[
\mathrm{KL}(p\|p_S)
\ge\delta_c\|v\|_2^2\int_0^1(1-t)e^{-ts}\,dt
=\frac{\delta_c}{2}(s-1+e^{-s}).
\]
The inequality extends to $s=0$ by continuity. The function
$h(s)=s-1+e^{-s}$ is increasing on $[0,\infty)$, strictly increasing
away from zero, and convex, so its inverse on $[0,\infty)$ is
increasing and concave. Pointwise inversion followed by Jensen gives
\[
\mathbb E_c\|\Delta\widetilde z\|_2
\le\frac{\tau}{\sqrt2}\mathbb E_c h^{-1}(2\mathrm{KL}_x/\delta_c)
\le\frac{\tau}{\sqrt2}h^{-1}(2\varepsilon_{\mathrm{KL}}(c)/\delta_c).
\]
Finally, $h(s)=s^2/2+O(s^3)$ near zero gives the stated small-error
behavior. This argument averages scalar bounds after applying the
input-specific Fisher inequality.
\end{proof}

For completeness, the reverse direction, which bounds the divergence by
the centered-logit error without any floor, follows from
$\|F_p\|_{\mathrm{op}}\le\max_i2p_i(1-p_i)\le1/2$ by
Gershgorin's theorem. Substituting this upper bound into
Eq.~\ref{eq:kl-integral} gives
$\mathrm{KL}(p_T\|p_S)\le\|\Delta\widetilde z\|_2^2/(4\tau^2)$.

\paragraph{Computing the bound.}
The inverse is a monotone one-dimensional root solve; define
$h^{-1}(0)=0$. A useful explicit upper bound is
$h^{-1}(a)\le a+\sqrt{a^2+2a}$ for $a\ge0$.
Indeed, $h'(s)=1-e^{-s}\ge s/(1+s)$ and hence
$h(s)\ge s^2/[2(1+s)]$ by integrating the lower bound
$t/(1+s)$ over $0\le t\le s$. This also shows that the certificate
remains finite for every finite KL, although a small teacher floor
can make it conservative.

\begin{corollary}[Input-specific teacher-floor KL bound]
\label{cor:pointwise-kl-floor}
On each evaluated input let
$\delta(x)=\min_i\mathrm{softmax}(z_T^{\mathrm{serve}}(x)/\tau)_i>0$
and let $\mathrm{KL}_x$ be the unscaled teacher-to-student KL.
Then
\begin{equation}
d_c\le J_c^{\mathrm{point}}:=
\frac{\tau}{\sqrt2}\mathbb E_c
h^{-1}\!\left(\frac{2\mathrm{KL}_x}{\delta(x)}\right)
\le\frac{\tau}{\sqrt2}
h^{-1}\!\left(\frac{2\mathbb E_c\mathrm{KL}_x}{\delta_c}\right)
\label{eq:pointwise-kl-floor}
\end{equation}
for any $0<\delta_c\le\min_{x:y=c}\delta(x)$.
Thus any certified classwise KL bound may replace $\rho_c$ or $d_c$
in the rowwise errors. When these quantities are directly measured,
their minimum remains a valid upper bound for $\rho_c$.
\end{corollary}

\begin{proof}
Apply the pointwise step in the proof of Corollary~\ref{cor:kl-floor}
using the input's own floor, then average. The second inequality uses
$\delta(x)\ge\delta_c$, monotonicity of $h^{-1}$, and its concavity.
The same actual audit temperature and served teacher must be used
throughout.
\end{proof}

\subsection{Conditional Payload Tests}
\label{app:bertest-cc-proof}

\begin{proof}[Proof of Theorem~\ref{thm:bertest-cc}]
Conditional on the declared information, the score vector and its
decoded signs are fixed. Each uniform independent target bit matches
its decoded sign with probability one half, proving the binomial
claim even when the score coordinates are dependent.

For the weighted test, put $a_k=s_k/\|s\|_2$, so $\sum_k a_k^2=1$.
For every $\lambda>0$, independence and
$\cosh(u)\le\exp(u^2/2)$ give
\[
\mathbb E\exp\!\left(\lambda\sum_k a_kb_k\right)
=\prod_k\cosh(\lambda a_k)\le\exp(\lambda^2/2).
\]
Chernoff's bound, minimized at $\lambda=t>0$, yields
$\Pr[\sum_k a_kb_k\ge t]\le e^{-t^2/2}$. Inverting this bound proves
super-uniformity of $p_{\mathrm{wt}}$. For $t\le0$, or a zero score
vector, the reported value one cannot cause rejection.

Finally, $\Pr[\min_jp_j\le a/q]\le\sum_j\Pr[p_j\le a/q]\le a$.
This conditional bound also holds after averaging over the conditioned
information. If at least $r$ bits are certified correct, then $M\ge r$;
monotonicity of the binomial upper tail gives the sufficient rejection condition.
\end{proof}

\paragraph{Executed decision rule and its assumptions.}
The score construction uses the suspect outputs, fixed carrier
indices or evaluation labels, and non-payload readout material.
The target $b$ is consulted only when computing bit agreement and
weighted agreement. Codebook conditioning is screened using $C$
alone, before suspect evaluation, rather than using successful
payload recovery. Validity still requires independence of the
suspect from the committed payload under the null; adaptive key
selection after inspecting a suspect is not covered.

We fix the available channels before evaluation and allocate a total
level $10^{-3}$ equally among two tests per available channel.
This gives per-test levels $5\times10^{-4}$ for one channel and
$2.5\times10^{-4}$ for both. The binomial calculation is analytic;
the weighted calculation uses the displayed conservative bound.
Neither requires a Monte Carlo rank approximation. Repeated keys on
one negative model assess conditional key behavior, whereas
independently trained negative models assess variation across
models; their experimental units are reported separately.

\paragraph{A conditional rank test for a scalar key score.}
\label{app:key-rank}
For a detector without a bit payload, a separate key-randomization
argument can calibrate a scalar statistic. Fix the suspect, query
selection, and the complete scoring procedure. Under a null for
which the observed key and $B$ fresh reference keys are exchangeable
and independent of these fixed quantities, their scores
$S_0,\ldots,S_B$ are exchangeable. Then
\[
p_{\mathrm{rank}}=
\frac{1+\sum_{j=1}^{B}\mathbf1\{S_j\ge S_0\}}{B+1}
\]
is super-uniform. To prove this, condition on the unordered scores.
At most $\lfloor a(B+1)\rfloor$ positions can have an upper-tail
rank no larger than $a(B+1)$; exchangeability makes the observed
position uniform. Counting ties against rejection gives
$\Pr(p_{\mathrm{rank}}\le a)\le a$.
The same fixed lowest score may represent every undefined statistic;
this preserves exchangeability and assigns an undefined observed
score $p_{\mathrm{rank}}=1$. With $B=1999$, the resolution is
$1/2000$. The probability is over the observed and reference-key
draws jointly, conditional on the fixed suspect and query procedure;
it need not remain bounded by the same level after conditioning on
an arbitrary realized reference bank. This argument does not
establish an empirical false-positive rate across a
population of trained models. It also does not justify choosing
queries, keys, or scoring settings after inspecting their scores.

\paragraph{Full-key randomization.}
\label{app:full-key}
Theorem~\ref{thm:bertest-cc} is exact when the payload is independent of
all other readout material. Our implementation expands one 32-byte seed
deterministically, and domain separation does not make the expanded
objects exactly independent. The rank argument above calibrates the
implemented detector without this assumption. For a key $\kappa$ and the
declared test family $\mathcal J$ of one access scenario, let
\[
T(\kappa;S)=\max_{j\in\mathcal J}\bigl[-\log p_j(\kappa;S)\bigr],
\]
where the analytic p-values of Theorem~\ref{thm:bertest-cc} serve only as
scores. Each of $B=1{,}999$ reference keys, drawn like the owner key
$\kappa_0$, regenerates the complete readout, including the codebook
selection, and
\[
p_{\mathrm{key}}=\frac{1+\sum_{i=1}^{B}\mathbf1\{T(\kappa_i;S)\ge T(\kappa_0;S)\}}{B+1}.
\]
\begin{proposition}[Full-key randomization]
\label{prop:full-key}
Fix the suspect, the audit inputs and the scoring procedure. If, under the
null, the owner and reference keys are exchangeable and independent of
these fixed quantities, then $p_{\mathrm{key}}$ is super-uniform, whatever
the deterministic dependence among the objects generated from one key.
\end{proposition}
\begin{proof}
The scores $T(\kappa_0;S),\ldots,T(\kappa_B;S)$ are exchangeable, so the
argument for $p_{\mathrm{rank}}$ applies, with ties counted against
rejection.
\end{proof}
The maximum already accounts for selection within $\mathcal J$, so
$p_{\mathrm{key}}$ is not multiplied by $|\mathcal J|$, and access
scenarios are never combined by an OR. The test reuses the saved suspect
outputs and needs no new query.

\subsection{Rank, Concentration, and Decoder Conditioning}
\label{app:capacity-cc-proof}

\begin{proof}[Proof of Theorem~\ref{thm:capacity-cc}]
Each class block has rank at most $\mathrm{rank}(C)$, proving the
deterministic rank bound. Let $D_c=\mathrm{diag}(\psi_c)$.
Since $C$ has unit columns, $H=C^{\!\top}C-I_K$ has zero diagonal and
\[
A^{\!\top}A-mI_K=\sum_{c=1}^m X_c,\qquad X_c=D_cHD_c.
\]
The $X_c$ are independent self-adjoint matrices with zero mean and
$\|X_c\|_{\mathrm{op}}=\|H\|_{\mathrm{op}}$. Because $D_c^2=I_K$,
\[
\mathbb E X_c^2=\mathbb E[D_cH^2D_c]
=\mathrm{diag}((H^2)_{11},\ldots,(H^2)_{KK}),\qquad
\left\|\sum_c\mathbb E X_c^2\right\|_{\mathrm{op}}=mv.
\]
Matrix Bernstein applied to both signs of the sum
\citep{Tropp2012} therefore gives
\[
\Pr\!\left(\left\|\sum_cX_c\right\|_{\mathrm{op}}\ge t\right)
\le2K\exp\!\left(-\frac{t^2}{2(mv+\|H\|_{\mathrm{op}}t/3)}\right).
\]
Substitution of $t=t_\beta$ bounds this probability by $\beta$.
If $H=0$, the Gram matrix is deterministically $mI_K$ and no
tail argument is needed. On the stated event,
$\lambda_{\min}(A^{\!\top}A)\ge m-t_\beta$.
When positive, this establishes full column rank and the singular-value
bound. In particular $t_\beta\le\theta m$, $0\le\theta<1$, gives
$\sigma_{\min}(A)\ge\sqrt{(1-\theta)m}$.
\end{proof}

\begin{proof}[Proof of Corollary~\ref{cor:capacity-cond}]
Full column rank gives $\widetilde b=b+A^+\eta$. Thus
\[
\|A^+\eta\|_\infty\le\|A^+\eta\|_2
\le\sigma_{\min}(A)^{-1}\|\eta\|_2
\]
and, independently,
$\|A^+\eta\|_\infty\le\|A^+\|_{\infty\to\infty}\|\eta\|_\infty$.
For every coordinate, $b_k\widetilde b_k
=1+b_k(A^+\eta)_k>0$ if the corresponding error is less than one.
Finally $\|\eta\|_2\le\sqrt{mP}\|\eta\|_\infty$ gives the
spectral-only sufficient condition.
\end{proof}

\paragraph{Interpreting the dimension bound.}
At $K=1024$, $m=100$, and $P=32$, the stacked matrix has $3200$
rows, so full column rank is dimensionally possible. Its actual
conditioning is a separate property. High-payload BER and bit-vote
significance measure empirical decoding and detection; they do not
measure $\sigma_{\min}(A)$ or establish the concentration condition.

\section{Training and Verification Details}
\label{app:training}

\paragraph{Key expansion.}
SHAKE128 expands the 32-byte owner key under a separate domain tag for
each keyed object. Bits map $0\mapsto-1$ and $1\mapsto+1$, and Gaussian
entries come from Box--Muller transforms of 32-bit uniforms. Payload bits
and class signatures use separate domains. Separation adds no entropy, so
the independent payload of Theorem~\ref{thm:bertest-cc} and the independent
signs of Theorem~\ref{thm:capacity-cc} idealize this expansion
(Appendix~\ref{app:full-key}). Of five Gaussian codebook
candidates with unit-norm columns, we keep the one minimizing
$\max_k[(K-1)^{-1}\sum_{j\ne k}\langle c_k,c_j\rangle^2]^{1/2}$, a choice
that depends on the key alone. The logit frame is a keyed Gaussian matrix,
centered and orthonormalized by QR decomposition, and its realized value
is stored with the model. Its columns lie in the nonzero eigenspace of the
multinomial Fisher matrix at the uniform reference distribution.

\paragraph{Teacher training.}
The teacher minimizes Eq.~\ref{eq:LT} with
$\mathcal L_{\mathrm{orth}}=\mathbb E\|G^{\!\top}z_T^{\mathrm{base}}\|_2^2$.
The auxiliary loss uses $Q$ keyed training anchors $a_j$. Let
$\eta_\ell=(u_T(x_\ell)^{\!\top}u_T(a_j))_{j=1}^Q$, let $B_k$ be keyed
symmetric matrices of unit operator norm, and let
$\phi_{\ell k}\in\{\pm1\}$ be keyed carrier targets independent of the
payload in the idealized key model. Then
\[
\mathcal L_{\mathrm{aux}}=\frac1{LK}\sum_{\ell,k}
\left[\tanh(\gamma\eta_\ell^{\!\top}B_k\eta_\ell)
-b_k\phi_{\ell k}\right]^2.
\]
Here $\gamma$ is a slope. This term shapes the same comparison space as the
second-moment hinge and adds no verification channel. Anchor features are
computed without augmentation or gradient and refreshed each epoch,
whereas carrier features carry gradients at every watermark step. The
watermark losses are evaluated every fourth minibatch on CIFAR and every
eighth on MiniImageNet. The hinge starts after three epochs, while
$\alpha$ and $\lambda_{\mathrm{orth}}$ ramp up over the first three.
Verification uses evaluation-mode features normalized as
$v/(\|v\|_2+10^{-8})$.

CIFAR-100 teachers train for 60 epochs with stochastic gradient descent
(SGD), using Nesterov momentum $0.9$, learning rate $0.06$, weight decay
$0.000954$, batch 64, gradient-norm clipping at one, eight linear warm-up
epochs and cosine decay to $0.01$ of the peak rate. The joint writer uses $\alpha=0.1247$,
$\lambda_{\mathrm{aux}}=2.143$, $\lambda_{\mathrm{sm}}=1.50881$,
$\lambda_{\mathrm{orth}}=1.23$, $\gamma=2.880$,
$m_{\mathrm{target}}=1.055107$, readout rank $r=16$, $Q=8$ anchors and
$L=256$ carriers. This target exceeds every attainable signed score, at
most one, so every hinge stays active and $\mathcal L_{\mathrm{sm}}$ reduces
to maximizing the average signed score. The single-writer arms set
$\alpha=\lambda_{\mathrm{orth}}=0$ or
$\lambda_{\mathrm{aux}}=\lambda_{\mathrm{sm}}=0$, and the ablation sets
$\lambda_{\mathrm{aux}}=0$.

\paragraph{Students.}
Students share each seed's initial weights with their teacher but copy no
trained weight. They train for 30 epochs with SGD (Nesterov momentum
$0.9$, learning rate $0.05$, weight decay $0.0005$, batch 128, clipping at
one) and cosine decay to $0.01$ of the initial rate. KL-KD minimizes
$0.59\,\mathcal L_{\mathrm{KD}}+0.41\,\mathcal L_{\mathrm{CE}}$ and FM-KD
minimizes $0.59(6.61)\,\mathcal L_{\mathrm{FD}}+0.41\,\mathcal L_{\mathrm{CE}}$,
with $\tau=4.538$ and $\mathcal L_{\mathrm{CE}}$ the cross-entropy. FM-KD
trains the student's comparison head, whereas KL-KD leaves it at
initialization. All results use the final epoch, without checkpoint
selection. Sharing the teacher's initialization is a matched control,
not a requirement: the students of Appendix~\ref{sec:crossarch-extension}
share no initial weights with their teachers and still inherit the mark
on the imitated channel in 39 of 40 cases.

\paragraph{Models and data.}
ResNet-18 exposes its 512-dimensional pooled feature, which an affine
comparison head maps to 128 dimensions before normalization, while an
independent linear classifier produces the logits. Cross-entropy reads the
pooled feature, so only the watermark losses train a teacher's comparison
head, and a clean teacher keeps its initial head up to scale. The other students map
their pooled features (ResNet-50: 2048, VGG-16-BN: 512, MobileNetV3-Large:
960) or ViT-S/16 class token (384) to the same 128-dimensional space.
Training views use random crops and horizontal flips on CIFAR, and
$84\times84$ random resized crops, flips and color jitter on MiniImageNet.
Evaluation and carrier views use no augmentation.

\paragraph{Protocol.}
The seeds are 42, 137, 271, 314, 1729, 2718, 3141, 5772, 6561 and 9999,
each with one unscreened 32-byte owner key, and clean models never read a
key. Recipes, attacks, data splits and tests were fixed before the
reported cohorts, without new tuning. The inherited recipe had been
selected on evaluation accuracy during development, so all comparisons
are fixed-recipe comparisons, and we do not regard the test data as
untouched.

\paragraph{Cost and queries.}
Marked CIFAR-100 teachers add 50,176 carrier-image forwards per epoch (256
carriers at every fourth of 782 minibatches) to 50,000 task-image
forwards. On five seed pairs run on identical graphics processing units, joint training takes
$2.29\pm0.04$ times as long as clean training. A feature audit queries the
256 carriers, a logit audit the 10,000 labeled test inputs (100 per
class), and dual access both. Null-key batteries reuse these outputs
without new queries.

\subsection{Applying the Certificates}
\label{app:design-checklist}
All certificate quantities are evaluated on saved outputs for the audit
inputs. For the feature channel, compute the teacher's signed margins and
$e_k^\Sigma$ after any declared alignment. For the logit channel, compute
$B_c$, $E_c$ and either $d_c$ or $(\delta_c,\varepsilon_{\mathrm{KL}}(c))$
on the class-stratified set, with the probability floor measured on the
served teacher at the audit temperature. Then check the rank and smallest
singular value of $C$ or $A$, and apply the per-class or stacked bound.
The tests, their level allocation and any key screening for conditioning
are fixed before querying the suspect. We compute in float64 with
\texttt{numpy.linalg.pinv} at \texttt{rcond}$=10^{-12}$, treating
$\sigma_i$ as nonzero when $\sigma_i>10^{-12}\sigma_1$.

\section{Supporting Experimental Results}
\label{app:additional_results}

Each subsection supports one claim of Section~\ref{sec:experiments}, in
the same order.

\subsection{Inheritance Follows the Imitated Surface}
\label{sec:nonvacuity}
\label{sec:component-details}

Table~\ref{tab:components-all-channels} reports both readouts of every arm
of Table~\ref{tab:components}. In every arm and under either attack, the
channel that the attack does not imitate stays at chance (48--50\% BER, no
detection). A mark is therefore inherited only through the imitated
surface, and an audit of one channel cannot stand in for the other. Adding
the logit writer after training reaches only 36.41\% logit BER and two of
ten KL-KD detections, against 11.09\% and ten of ten for joint training.
Both tests reject for all twenty joint students on the imitated channel.
Without the auxiliary carrier loss, feature BER rises to 7.66\% while
logit BER slightly falls.

\paragraph{Shared versus independent payloads.}
Under the independent-sign model, the class code $\omega_c=\psi_c\odot b$
is uniform and independent of $b$. A logit writer with an independent
payload $b'$ therefore equals a shared-payload writer with signatures
$\psi_c\odot b\odot b'$, and sharing the payload changes no distribution.
A teacher whose logit payload is drawn from a hash of the key under a
separate tag confirms this on the first five seeds, with each channel
scored against its own payload. Its FM-KD students keep 3.75\% feature BER
and 55.2 certified bits (shared: 3.75\% and 54.0), and its KL-KD students
reach 6.25\% logit BER (shared: 9.69\%), a gap that reflects a different
signature draw and training run. Every student is detected under its
matching single access and under dual access, and student accuracies
differ by at most 0.61 points.
Joint training changes paired student accuracy by $-0.04\pm0.40$ points
after FM-KD and $-0.39\pm0.47$ after KL-KD (mean and SD of ten within-seed
differences).

\begin{table}[t]
\centering\scriptsize
\caption{Both readouts of every component arm (CIFAR-100, ten seeds). F/L BER: feature/logit bit error rate in \% (mean$_{\pm\mathrm{SD}}$). F/L det.: detections under feature-only or logit-only access. Dual det.: detections with both channels exposed (four tests at total level $10^{-3}$).}
\label{tab:components-all-channels}
\begin{tabular}{llrrrrr}
\toprule
Method & Attack & F BER & F det. & L BER & L det. & Dual det. \\
\midrule
Feature writer only & FM-KD & $2.66_{\pm 3.46}$ & 10/10 & $48.91_{\pm 5.99}$ & 0/10 & 10/10 \\
Feature writer only & KL-KD & $48.12_{\pm 7.82}$ & 0/10 & $50.00_{\pm 7.37}$ & 0/10 & 0/10 \\
Logit writer only & FM-KD & $50.00_{\pm 6.55}$ & 0/10 & $49.69_{\pm 4.53}$ & 0/10 & 0/10 \\
Logit writer only & KL-KD & $48.91_{\pm 7.29}$ & 0/10 & $11.41_{\pm 4.04}$ & 10/10 & 10/10 \\
Feature + posthoc logits & FM-KD & $2.66_{\pm 3.46}$ & 10/10 & $48.91_{\pm 5.99}$ & 0/10 & 10/10 \\
Feature + posthoc logits & KL-KD & $49.06_{\pm 7.73}$ & 0/10 & $36.41_{\pm 6.12}$ & 2/10 & 1/10 \\
Joint, no auxiliary loss & FM-KD & $7.66_{\pm 3.86}$ & 10/10 & $49.38_{\pm 5.95}$ & 0/10 & 10/10 \\
Joint, no auxiliary loss & KL-KD & $49.53_{\pm 6.67}$ & 0/10 & $9.53_{\pm 4.06}$ & 10/10 & 10/10 \\
TwinMark (joint) & FM-KD & $3.44_{\pm 2.74}$ & 10/10 & $49.53_{\pm 5.56}$ & 0/10 & 10/10 \\
TwinMark (joint) & KL-KD & $49.84_{\pm 7.49}$ & 0/10 & $11.09_{\pm 5.02}$ & 10/10 & 10/10 \\
\bottomrule
\end{tabular}
\end{table}

\paragraph{Utility of the verified embedding.}
The verifier reads the comparison embedding, which the classifier does not
use. A fixed cosine 5-nearest-neighbor (5-NN) classifier measures it
directly, with a gallery of the first 100 non-carrier training images per
class and the 10,000 test images as queries
(Table~\ref{tab:verified-utility}). Marking lowers 5-NN accuracy in every
seed, by $4.36\pm0.60$ points in FM-KD students, whose classifier accuracy
moves by only $-0.04$ points, and by 1.91 points in teachers. The mark does not collapse
the embedding. With $\mu$ and $\Sigma$ the test-set mean and second
moment, marked FM-KD students have a higher participation rank
$r_{\mathrm{part}}=(\mathrm{tr}\,\Sigma)^2/\mathrm{tr}(\Sigma^2)$ than clean
ones, marked models have a far lower mean-energy fraction
$f_{\mathrm{mean}}=\|\mu\|_2^2/\mathrm{tr}\,\Sigma$, and the
centered term $\mathrm{tr}(W_k\mathrm{Cov}(u))$ carries about 96\% of the
signed carrier scores.

\begin{table}[h]
\centering\small
\caption{The verified 128-dimensional embedding (CIFAR-100, ten seeds,
mean$\pm$SD). Accuracies in \%. Teacher classifier accuracy uses the base head.}
\label{tab:verified-utility}
\begin{tabular}{lcccc}
\toprule
Model & Classifier acc. & 5-NN acc. & $r_{\mathrm{part}}$ & $f_{\mathrm{mean}}$ \\
\midrule
Clean teacher & $62.64\pm0.29$ & $61.66\pm0.19$ & $11.73\pm1.18$ & $0.242\pm0.019$ \\
Marked teacher & $61.36\pm0.38$ & $59.76\pm0.37$ & $11.64\pm0.57$ & $0.025\pm0.007$ \\
Clean-teacher FM-KD student & $58.19\pm0.27$ & $54.39\pm0.34$ & $7.66\pm0.81$ & $0.319\pm0.022$ \\
Marked-teacher FM-KD student & $58.15\pm0.36$ & $50.03\pm0.43$ & $11.53\pm0.71$ & $0.040\pm0.010$ \\
\bottomrule
\end{tabular}
\end{table}

\subsection{A Matched Distillation Watermark}
\label{sec:coswm-comparison}

CosWM \citep{charette2022cosine} writes a cosine signal into the served
probabilities and detects it spectrally. We train CosWM teachers with the
published writer ($\epsilon=0.05$, angular frequency 30, target class 0,
one private random pixel projection per seed) under the data,
architectures, initializations and budgets of Table~\ref{tab:components}.
Their students are distilled by FM-KD and by KL-KD on the served
log-probabilities. The native detector computes a Lomb--Scargle
SNR on 200 class-0 training images and detects
when SNR$>8$. It flags nine of ten teachers, no student and no clean
teacher (Table~\ref{tab:coswm-matched}).

\begin{table}[t]
\caption{CosWM under the matched protocol (CIFAR-100, ten seeds, mean$\pm$SD). Clean and CosWM acc.: accuracy in \% of the clean and CosWM lineages. Native SNR: CosWM's spectral statistic. Detected: models with SNR$>8$, the published rule, which flags none of the ten clean teachers.}
\label{tab:coswm-matched}
\centering\small
\begin{tabular}{lrrrr}
\toprule
Stage & Clean acc. (\%) & CosWM acc. (\%) & Native SNR & Detected \\
\midrule
Teacher & $62.64\pm0.29$ & $62.44\pm0.22$ & $12.16\pm2.98$ & 9/10 \\
Logit KD & $59.51\pm0.38$ & $59.38\pm0.29$ & $1.22\pm1.04$ & 0/10 \\
Feature KD & $58.19\pm0.27$ & $57.91\pm0.23$ & $0.66\pm0.62$ & 0/10 \\
\bottomrule
\end{tabular}

\end{table}

The native rule has no calibrated level, so we also audit both methods
with 100 held-out queries at conditional level $10^{-3}$. TwinMark splits
the level over its two logit tests on one input per class. CosWM keeps its
statistic on the 100 test inputs of class 0 but replaces the threshold by
a rank test against 1,999 fresh reference keys. With owner statistic $S_0$
and reference statistics $S_j$, it rejects when
\[
 p_{\mathrm{rank}}=
 \frac{1+\sum_{j=1}^{1999}\mathbf{1}\{S_j\ge S_0\}}{2000}
 \le 10^{-3}.
\]
This test is valid when the owner and reference keys are exchangeable
(Appendix~\ref{app:key-rank}). TwinMark then detects nine of ten KL-KD
students and CosWM none. Both miss the FM-KD students, whose mark lives in
features, and neither flags a clean model (Table~\ref{tab:common-query}).
Among students of detected teachers, TwinMark detects nine of ten and CosWM
none of six. CosWM reports an ownership signal rather than a payload, and its
lower teacher detectability limits conclusions about relative robustness
to distillation.

\begin{table}[t]
\centering\small
\caption{Logit audits with 100 test queries at total conditional level $10^{-3}$ (ten seeds). Accuracy: full-test-set accuracy in \% of the marked models (mean$\pm$SD). Detected: marked models detected. Clean flags: shared clean models flagged. TwinMark queries one input per class, CosWM all 100 test inputs of its target class.}
\label{tab:common-query}
\begin{tabular}{lrrrrrr}
\toprule
& \multicolumn{2}{c}{Accuracy (\%)} & \multicolumn{2}{c}{Detected} & \multicolumn{2}{c}{Clean flags} \\
Stage & TwinMark & CosWM & TwinMark & CosWM & TwinMark & CosWM \\
\midrule
Teacher & $61.35\pm0.41$ & $62.44\pm0.22$ & 10/10 & 6/10 & 0/10 & 0/10 \\
KL-KD & $59.12\pm0.36$ & $59.38\pm0.29$ & 9/10 & 0/10 & 0/10 & 0/10 \\
FM-KD & $58.15\pm0.36$ & $57.91\pm0.23$ & 0/10 & 0/10 & 0/10 & 0/10 \\
\bottomrule
\end{tabular}
\end{table}

\subsection{Certificates and the Key-Only Target}
\label{sec:certificates-key-only}

Table~\ref{tab:certificate-coverage} evaluates the five sufficient bounds
on the saved outputs of the ten joint students. Only the second-moment
bound certifies bits, 50 to 60 of 64 per FM-KD student (50, 60, 54, 53,
53, 54, 56, 54, 57 and 53 in seed order). At total level $10^{-3}$, 46
certified bits force the analytic vote test of the two-test feature family
to reject, and 47 the four-test dual family. It thus certifies detection in
every student, though not complete recovery, on the fixed carriers.

\begin{table}[t]
\centering\small
\caption{Certificate coverage for the ten joint students (feature bounds after FM-KD, logit bounds after KL-KD). Certified bits: bits certified out of 64 (mean$_{\pm\mathrm{SD}}$ and range over seeds). Detection certified: students whose certified bits force the analytic vote test of Theorem~\ref{thm:bertest-cc} to reject (46 bits in the two-test family). Full-key randomization is evaluated separately (Appendix~\ref{app:full-key}).}
\label{tab:certificate-coverage}
\begin{tabular}{lrrr}
\toprule
Bound & Certified bits /64 & Range & Detection certified \\
\midrule
Feature: norm bound & $0.00_{\pm 0.00}$ & 0--0 & 0/10 \\
Feature: directional & $0.00_{\pm 0.00}$ & 0--0 & 0/10 \\
Feature: second moment & $54.40_{\pm 2.72}$ & 50--60 & 10/10 \\
Logit: rowwise transfer & $0.00_{\pm 0.00}$ & 0--0 & 0/10 \\
Logit: input-specific KL & $0.00_{\pm 0.00}$ & 0--0 & 0/10 \\
\bottomrule
\end{tabular}
\end{table}

Table~\ref{tab:logit-diagnostics} reports the logit bounds. The stacked
design has full rank, with $\sigma_{\min}(A)$ between 8.589 and 8.743, and
every teacher margin is positive (at least 0.511), yet the largest rowwise
bound (10.66 to 11.28) exceeds the smallest margin about twentyfold, and
the teacher-floor bound is looser. Good conditioning rules out spectral
amplification but does not locate the slack. Table~\ref{tab:logit-slack}
therefore relaxes the observed change one inequality at a time. With
$g_c=G^{\!\top}(\bar z_S(c)-\bar z_T^{\mathrm{serve}}(c))$ and $D_{k,c}$ the
class-$c$ block of row $k$ of $A^+$, the stages are $E_0=\alpha^{-1}|\sum_cD_{k,c}g_c|$,
$E_1=\alpha^{-1}\sum_c|D_{k,c}g_c|$, which drops cross-class cancellation,
the blockwise bound $E_2=\alpha^{-1}\sum_c\|D_{k,c}\|_2\|g_c\|_2$, the
centered bound $E_3$ of Theorem~\ref{thm:klcert-cc}, and the KL bounds of
Corollary~\ref{cor:kl-floor} with pointwise ($E_4^{\mathrm p}$) or class
($E_4^{\mathrm c}$) floors. Each stage bounds the previous one for every
bit. The observed changes would certify 569 of 640 bits, but $E_1$
certifies none, so certification is lost with cross-class cancellation,
before any KL step. The KL conversion contributes a factor of about 3.5
to the 480-fold slack from $E_0$ to $E_4^{\mathrm c}$. In every student, 61 to 64
of the 64 signed score changes shrink the margin.

\begin{table}[h]
\centering\small
\caption{Stagewise slack of the logit bounds for the ten joint KL-KD
students (640 bits). Certified: bits with $\mu_k>E_{j,k}$. Ratio: median
over students of the median $E_{j,k}/\mu_k$.}
\label{tab:logit-slack}
\begin{tabular}{lcccccc}
\toprule
Stage & $E_0$ & $E_1$ & $E_2$ & $E_3$ & $E_4^{\mathrm p}$ & $E_4^{\mathrm c}$ \\
\midrule
Certified & 569 & 0 & 0 & 0 & 0 & 0 \\
Ratio & 0.56 & 2.56 & 17.7 & 76.1 & 213 & 269 \\
\bottomrule
\end{tabular}
\end{table}

\begin{table}[t]
\centering\small
\caption{Logit-certificate diagnostics for the ten joint KL-KD students ($K=64$, $m=100$, $P=32$, $\tau=4.538$). Each statistic is computed per seed and summarized by its mean$\pm$SD and range over seeds.}
\label{tab:logit-diagnostics}
\begin{tabular}{lcc}
\toprule
Per-seed statistic & Mean $\pm$ SD & Range \\
\midrule
Smallest stacked-design singular value & $8.669 \pm 0.0567$ & $8.589$--$8.743$ \\
Smallest signed teacher logit margin & $0.5334 \pm 0.0153$ & $0.511$--$0.5586$ \\
Largest rowwise transfer bound & $10.96 \pm 0.214$ & $10.66$--$11.28$ \\
Smallest served-teacher probability & $0.003072 \pm 0.000292$ & $0.002471$--$0.003374$ \\
Mean unscaled KL divergence & $0.01167 \pm 0.000257$ & $0.01128$--$0.01219$ \\
Largest classwise centered-logit error & $7.43 \pm 0.132$ & $7.139$--$7.54$ \\
Largest input-specific KL class bound & $24.27 \pm 0.893$ & $23.48$--$26.49$ \\
Largest base-logit residual & $0.1437 \pm 0.012$ & $0.1284$--$0.1625$ \\
Largest class-conditioning residual & $0.9163 \pm 0.0883$ & $0.8055$--$1.119$ \\
\bottomrule
\end{tabular}
\end{table}

\paragraph{Key-only feasible target.}
\label{sec:target-control}
The hinge target can be fixed from the key before training. Let
$H_k=b_kW_k$ and define
\[
t_\star=\max_{\Sigma\succeq0,\,\mathrm{tr}(\Sigma)=1}
\min_k\mathrm{tr}(H_k\Sigma).
\]
Any trace-one positive semidefinite (PSD) witness $\Sigma_0$ gives the
lower bound $\ell=\min_k\mathrm{tr}(H_k\Sigma_0)\le t_\star$. For any
simplex weights $q_k\ge0$, $\sum_kq_k=1$,
\[
t_\star\le\lambda_{\max}\!\left(\sum_kq_kH_k\right).
\]
Indeed, the minimum of the signed scores is no greater than their
$q$-weighted average, and the maximum of that average over trace-one PSD
matrices is the displayed eigenvalue. We build $\Sigma_0$ by column
generation over mixtures of $I_d/d$ and rank-one matrices, in which a
linear program sets the mixture and its dual weights propose the next
leading eigenvector (float64, at most 512 iterations, bracket gap
$10^{-5}$). If $\ell>\eta=10^{-8}$, we set
$m_{\mathrm{target}}=(\ell-\eta)/2$, so $0<m_{\mathrm{target}}<\ell$ and
$\Sigma_0$ attains every signed score at least $\ell$. If $\ell\le\eta$,
the procedure reports that no positive target is certified, without
inferring that zero is feasible or resampling the key. No data or model
outcome enters this choice. Feasibility refers to the trace-one PSD
relaxation, and a trained network need not realize the witness. All ten
keys give $\ell>\eta$, with targets from 0.03036 to 0.04328, and every
resulting teacher has a positive minimum margin (0.02614 to 0.03886). All
ten FM-KD students then recover all 64 feature bits
(Table~\ref{tab:components}), with paired accuracy differences from the
primary target of $+0.01\pm0.27$ points after FM-KD and $-0.17\pm0.65$
after KL-KD. The second-moment bound certifies 34, 64, 63, 63, 53, 64, 63,
64, 64 and 58 bits, that is, complete recovery in four seeds and detection
in nine.

\subsection{Post-Processing and Evasion Attacks}
\label{sec:battery}

Nine fixed attacks act on the ten CIFAR-100 teachers and on the five
teacher pairs of CIFAR-10 and MiniImageNet. Five change the weights of a
stolen teacher: INT8 and INT4 quantization, 50\% magnitude pruning,
Gaussian weight noise ($\sigma=10^{-3}$) and ten epochs of fine-tuning.
Four transform the served embedding: a random rotation, a non-orthogonal
warp $I+0.1E/\sqrt d$ with Gaussian $E$, Gaussian feature noise
($\sigma=0.05$) and ZCA whitening. A
thief who holds the weights can drop the logit writer, which acts only at
serving time, so every attack is scored on the feature channel, with the
owner key, the feature-only test and the second-moment certificate.

\providecolor{twFeat}{HTML}{0072B2}
\providecolor{twFeatTint}{HTML}{E1EEF7}
\providecolor{twLogit}{HTML}{E69F00}
\providecolor{twLogitTint}{HTML}{FCEFD9}
\begin{table}[t]
\centering\scriptsize
\setlength{\tabcolsep}{2.4pt}
\renewcommand{\arraystretch}{1.06}
\caption{Post-processing attacks on stolen joint teachers. They test the feature channel because the logit writer acts at serving time. Acc.: base-head accuracy in \%. F BER: feature bit error rate in \% (mean$_{\pm\mathrm{SD}}$). Det.: feature-access detections at total level $10^{-3}$. Cert.: detections certified by the second-moment bound. No clean teacher is flagged under any attack (Table~\ref{tab:battery-detail}).}
\label{tab:battery-all}
\begin{tabular}{@{}l|rrcc|rrcc|rrcc@{}}
\toprule
 & \multicolumn{4}{c|}{\textbf{CIFAR-100} (10 seeds)} & \multicolumn{4}{c|}{\textbf{CIFAR-10} (5 seeds)} & \multicolumn{4}{c}{\textbf{MiniImageNet} (5 seeds)} \\
Attack on the teacher & Acc. & F BER & Det. & Cert. & Acc. & F BER & Det. & Cert. & Acc. & F BER & Det. & Cert. \\
\midrule
None & $61.36$ & $3.28_{\pm 3.33}$ & 10/10 & 10/10 & $87.16$ & $2.50_{\pm 4.07}$ & 5/5 & 5/5 & $60.05$ & $19.53_{\pm 4.82}$ & 5/5 & 5/5 \\
\multicolumn{13}{@{}l}{\textit{Weights of a stolen teacher}} \\
INT8 quantization & $61.36$ & $3.28_{\pm 3.33}$ & 10/10 & 10/10 & $87.14$ & $2.50_{\pm 4.07}$ & 5/5 & 5/5 & $59.94$ & $19.53_{\pm 4.82}$ & 5/5 & 5/5 \\
INT4 quantization & $51.04$ & $3.91_{\pm 2.78}$ & 10/10 & 10/10 & $82.92$ & $2.50_{\pm 2.61}$ & 5/5 & 5/5 & $48.12$ & $17.34_{\pm 3.92}$ & 5/5 & 0/5 \\
Pruning (50\%) & $57.57$ & $3.44_{\pm 2.83}$ & 10/10 & 10/10 & $84.15$ & $3.12_{\pm 3.83}$ & 5/5 & 5/5 & $54.26$ & $17.97_{\pm 5.58}$ & 5/5 & 0/5 \\
Weight noise & $60.15$ & $3.28_{\pm 2.90}$ & 10/10 & 10/10 & $85.99$ & $2.50_{\pm 4.07}$ & 5/5 & 5/5 & $55.14$ & $18.12_{\pm 5.25}$ & 5/5 & 1/5 \\
Fine-tuning (10 ep.) & $57.23$ & $3.12_{\pm 2.85}$ & 10/10 & 10/10 & $83.58$ & $2.50_{\pm 4.07}$ & 5/5 & 4/5 & $52.36$ & $17.19_{\pm 5.98}$ & 5/5 & 1/5 \\
\multicolumn{13}{@{}l}{\textit{Served embedding (logits unchanged)}} \\
Orthogonal rotation & $61.36$ & $51.41_{\pm 7.81}$ & 0/10 & 0/10 & $87.16$ & $52.50_{\pm 8.67}$ & 0/5 & 0/5 & $60.05$ & $50.00_{\pm 5.00}$ & 0/5 & 0/5 \\
Non-orthogonal warp & $61.36$ & $3.28_{\pm 3.72}$ & 10/10 & 10/10 & $87.16$ & $2.50_{\pm 4.07}$ & 5/5 & 5/5 & $60.05$ & $19.84_{\pm 4.34}$ & 5/5 & 2/5 \\
Feature noise & $61.36$ & $5.47_{\pm 2.47}$ & 10/10 & 10/10 & $87.16$ & $6.25_{\pm 5.85}$ & 5/5 & 1/5 & $60.05$ & $20.62_{\pm 4.90}$ & 5/5 & 3/5 \\
ZCA whitening & $61.36$ & $6.56_{\pm 2.64}$ & 10/10 & 0/10 & $87.16$ & $4.38_{\pm 6.48}$ & 5/5 & 0/5 & $60.05$ & $40.16_{\pm 6.78}$ & 3/5 & 0/5 \\
\bottomrule
\end{tabular}
\end{table}

On all three datasets, every weight attack, the warp and feature noise
keep detection in every joint teacher (Table~\ref{tab:battery-all}). ZCA
whitening misses two MiniImageNet teachers and removes every certificate.
The rotation hides the mark (51.41\% BER on CIFAR-100), as
Proposition~\ref{thm:rot} (Appendix~\ref{app:readout-definitions})
predicts. An owner-side Procrustes alignment restores detection but also
flags every clean teacher, because the fit uses the owner's
payload-bearing embeddings, so aligned scores cannot serve as a test
(Table~\ref{tab:battery-detail}). No clean teacher is flagged under any
attack. Certificates are thinner on the two smaller studies, most on
MiniImageNet, whose 128-bit payload starts at 19.53\% feature BER.

The same boundaries hold for extracted students. A fixed random rotation
of the ten marked FM-KD students' embeddings preserves every cosine, hence
5-NN accuracy (Appendix~\ref{sec:component-details}), yet leaves none
detected (48.6\% BER, no certified bit). Replacing every carrier embedding
by one key-independent unit vector $u_0$, so that
$\Sigma_C=u_0u_0^{\!\top}$, changes no other output and detects no
marked or clean student (smallest family bound 0.0081).

\begin{table}[t]
\centering\scriptsize
\setlength{\tabcolsep}{2.6pt}
\caption{Battery controls. Clean acc.: accuracy in \% of the attacked clean teachers. Flags: clean teachers detected with the owner key. Cos.: mean cosine between attacked and original carrier embeddings of the joint teachers. The owner-aligned rotation is fitted to the owner's embeddings, which breaks the null, so it is diagnostic only. The last row reads the served logits of the unattacked joint teachers.}
\label{tab:battery-detail}
\begin{tabular}{@{}l|rcr|rcr|rcr@{}}
\toprule
 & \multicolumn{3}{c|}{\textbf{CIFAR-100}} & \multicolumn{3}{c|}{\textbf{CIFAR-10}} & \multicolumn{3}{c}{\textbf{MiniImageNet}} \\
Attack & Clean acc. & Flags & Cos. & Clean acc. & Flags & Cos. & Clean acc. & Flags & Cos. \\
\midrule
None & $62.65_{\pm 0.29}$ & 0/10 & $1.000$ & $87.77_{\pm 0.25}$ & 0/5 & $1.000$ & $61.45_{\pm 0.61}$ & 0/5 & $1.000$ \\
INT8 quantization & $62.59_{\pm 0.34}$ & 0/10 & $0.999$ & $87.75_{\pm 0.21}$ & 0/5 & $1.000$ & $61.43_{\pm 0.62}$ & 0/5 & $1.000$ \\
INT4 quantization & $52.61_{\pm 3.03}$ & 0/10 & $0.766$ & $83.64_{\pm 0.45}$ & 0/5 & $0.883$ & $52.24_{\pm 1.06}$ & 0/5 & $0.828$ \\
Pruning (50\%) & $58.66_{\pm 0.76}$ & 0/10 & $0.875$ & $84.09_{\pm 1.20}$ & 0/5 & $0.856$ & $55.89_{\pm 0.93}$ & 0/5 & $0.946$ \\
Weight noise & $61.58_{\pm 0.35}$ & 0/10 & $0.953$ & $86.81_{\pm 0.42}$ & 0/5 & $0.948$ & $56.47_{\pm 1.12}$ & 0/5 & $0.955$ \\
Fine-tuning (10 ep.) & $57.89_{\pm 0.38}$ & 0/10 & $0.665$ & $84.00_{\pm 0.78}$ & 0/5 & $0.674$ & $53.65_{\pm 0.99}$ & 0/5 & $0.926$ \\
Orthogonal rotation & $62.65_{\pm 0.29}$ & 0/10 & $0.010$ & $87.77_{\pm 0.25}$ & 0/5 & $0.019$ & $61.45_{\pm 0.61}$ & 0/5 & $-0.004$ \\
Non-orthogonal warp & $62.65_{\pm 0.29}$ & 0/10 & $0.995$ & $87.77_{\pm 0.25}$ & 0/5 & $0.995$ & $61.45_{\pm 0.61}$ & 0/5 & $0.995$ \\
Feature noise & $62.65_{\pm 0.29}$ & 0/10 & $0.902$ & $87.77_{\pm 0.25}$ & 0/5 & $0.711$ & $61.45_{\pm 0.61}$ & 0/5 & $0.953$ \\
ZCA whitening & $62.65_{\pm 0.29}$ & 0/10 & $0.684$ & $87.77_{\pm 0.25}$ & 0/5 & $0.663$ & $61.45_{\pm 0.61}$ & 0/5 & $0.607$ \\
\midrule
Rotation, owner-aligned & \multicolumn{3}{c|}{joint 10/10, clean 10/10} & \multicolumn{3}{c|}{joint 5/5, clean 5/5} & \multicolumn{3}{c}{joint 5/5, clean 5/5} \\
None, served logits & \multicolumn{3}{c|}{L BER $0.00_{\pm 0.00}$, det. 10/10} & \multicolumn{3}{c|}{L BER $13.75_{\pm 9.78}$, det. 4/5} & \multicolumn{3}{c}{L BER $0.00_{\pm 0.00}$, det. 5/5} \\
\bottomrule
\end{tabular}
\end{table}

\subsection{Payload Size, Multiplexing and Additional Datasets}
\label{sec:capacity-extension}
\label{sec:classification-datasets}

The payload sweep changes only $K$ in the primary joint recipe, on the
first five seeds ($K=64$ reuses the primary models). It fixes the
per-coordinate amplitude $\alpha$, so the nominal injection energy
$\alpha^2K$ grows with $K$. The stacked logit
design has full column rank at every size, with the conditioning of
Table~\ref{tab:payload}. The second-moment bound certifies on average 30.6,
54.0, 76.6, 89.4, 95.8 and 90.8 feature bits for $K$ from 32 to 1,024. It
certifies detection in all five FM-KD students up to $K=64$, in one at
$K=128$ and in none beyond, and the logit bounds certify no bit at any
size, whereas the calibrated tests detect every student at every size.
The served teacher replaces labels by predictions, so its code
contribution has design $A_{\mathrm{serve}}=(Q\otimes I_P)A$, where
$Q_{cj}$ is the fraction of class-$c$ test inputs predicted as $j$. For
the ten primary teachers ($K=64$), $A_{\mathrm{serve}}$ keeps full rank,
with $\sigma_{\min}$ from 5.37 to 5.51 against 8.59 to 8.74 for $A$, and
every noiseless served margin $b_k(A^+A_{\mathrm{serve}}b)_k$ is positive (at
least 0.52).

\paragraph{Fixed energy and multiplexing.}
Table~\ref{tab:energy-mux} sets $\alpha_K=\alpha_{64}\sqrt{64/K}$, which
holds the nominal energy $\alpha^2K$ at its primary value, and adds a
non-multiplexed control with $\psi_c=\mathbf1$ at $K=128$, otherwise
identical. Teachers and KL-KD students use the primary recipe and seeds.
Mean serving energy stays within 6\% of nominal in every arm, and mean
student accuracy varies by under half a point. Logit BER rises from 1.2\% at $K=32$ to 40.5\% at
$K=1{,}024$, yet every student is detected. Removing multiplexing cuts the
rank to 32, which leaves 29.5\% of the bits wrong even without noise, and
raises student BER by $12.5\pm1.9$ points (in every seed), while both arms
stay detected. Multiplexing thus governs recovery rather than detection,
and neither sweep measures capacity at a fixed distortion.

\begin{table}[h]
\centering\footnotesize
\setlength{\tabcolsep}{3.5pt}
\caption{Fixed-energy sweep and multiplexing ablation (CIFAR-100, KL-KD,
first five seeds, mean$\pm$SD). Energy: served teacher's mean squared
logit perturbation (nominal 0.995). TV: mean total variation between served
and base probabilities. ECE: 15-bin expected calibration error of the
served teacher. TV and ECE in \%. The $K=64$ row is the primary cohort.}
\label{tab:energy-mux}
\begin{tabular}{llcrcccccc}
\toprule
Design & $K$ & $\alpha$ & Rank & Energy & TV & ECE & Student acc. & Logit BER & Exact/Det. \\
\midrule
Multiplexed & 32 & 0.1764 & 32 & 0.99 & 2.65 & 7.35 & $59.08\pm0.51$ & $1.2\pm1.7$ & 3/5, 5/5 \\
 & 64 & 0.1247 & 64 & 1.00 & 2.52 & 4.20 & $59.03\pm0.45$ & $9.7\pm2.0$ & 0/5, 5/5 \\
 & 128 & 0.0882 & 128 & 1.00 & 2.57 & 4.64 & $59.31\pm0.17$ & $17.7\pm2.1$ & 0/5, 5/5 \\
 & 256 & 0.0624 & 256 & 1.00 & 2.54 & 5.05 & $59.01\pm0.45$ & $24.7\pm2.1$ & 0/5, 5/5 \\
 & 512 & 0.0441 & 512 & 1.01 & 2.65 & 5.71 & $58.94\pm0.43$ & $33.5\pm1.4$ & 0/5, 5/5 \\
 & 1{,}024 & 0.0312 & 1{,}024 & 0.98 & 2.55 & 5.28 & $58.92\pm0.18$ & $40.5\pm2.0$ & 0/5, 5/5 \\
\midrule
Non-multiplexed & 128 & 0.0882 & 32 & 0.94 & 2.50 & 4.85 & $58.82\pm0.32$ & $30.2\pm1.4$ & 0/5, 5/5 \\
\bottomrule
\end{tabular}
\end{table}

CIFAR-10 ($K=32$, $P=2$) and a 100-class supervised MiniImageNet
\citep{VinyalsBLKW16} ($K=128$, $P=16$) repeat the protocol with ResNet-18
models on the first five seeds (Table~\ref{tab:classification-datasets}).
Carriers are the first 256 training images, which puts all 256 carriers in
class zero in MiniImageNet's class-sorted order. The MiniImageNet test view
keeps 4,998 images, excluding two train/test duplicates with conflicting
labels (original indices 3403 and 3439), and this cohort does not measure
clean detector flags. The unmatched channel again stays at chance
(45--56\% BER). Every MiniImageNet KL-KD student recovers all 128 bits,
more than its 100 classes. On CIFAR-10, FM-KD detects every student, but
KL-KD raises logit BER from the teachers' 13.75\% to 31.25\% without
detection. Its stacked decoder has rank 20 for 32 payload coordinates, the
only design whose payload exceeds its rank ceiling, so the full-rank guarantee does not
apply. Noiseless decoding of these designs still misses only 7.5\% of the
bits on average (0 to 18.75\% over the five keys), far below the students'
31.25\%, so rank alone does not explain the loss. The 2,000 new
CIFAR-10.1 images \citep{pmlr-v97-recht19a} repeat this pattern under the
logit test. Four of five joint teachers are detected (12.50\% BER),
neither KL-KD nor FM-KD students are detected, and no clean model is
flagged. These decisions are not merged with CIFAR-10 into an additional
cross-dataset OR.

\begin{table}[t]
\caption{CIFAR-10 ($K=32$, $P=2$) and MiniImageNet ($K=128$, $P=16$) with ResNet-18 models (five seeds, mean$\pm$SD in \%). Clean/Joint acc.: accuracy of the clean and marked lineages. $\Delta$ acc.: paired difference in points. Feature/Logit BER: marked models. Det.\ F/L/D: marked models detected under feature, logit and dual access at total level $10^{-3}$.}
\label{tab:classification-datasets}
\centering\small
\setlength{\tabcolsep}{3pt}
\begin{tabular}{lrrrrrc}
\toprule
Stage & Clean acc. & Joint acc. & $\Delta$ acc. & Feature BER & Logit BER & Det. F/L/D \\
\midrule
\multicolumn{7}{l}{CIFAR-10: $K=32$, $P=2$} \\
Teacher & $87.77\pm0.25$ & $87.19\pm0.19$ & $-0.59\pm0.30$ & $2.50\pm4.07$ & $13.75\pm9.78$ & 5/4/5 \\
KL-KD & $85.92\pm0.32$ & $85.63\pm0.14$ & $-0.29\pm0.26$ & $55.62\pm7.46$ & $31.25\pm5.85$ & 0/0/0 \\
FM-KD & $85.60\pm0.18$ & $85.55\pm0.33$ & $-0.05\pm0.35$ & $2.50\pm4.07$ & $47.50\pm7.78$ & 5/0/5 \\
\midrule
\multicolumn{7}{l}{Supervised MiniImageNet: $K=128$, $P=16$} \\
Teacher & $61.45\pm0.62$ & $59.51\pm0.52$ & $-1.94\pm0.93$ & $19.53\pm4.82$ & $0.00\pm0.00$ & 5/5/5 \\
KL-KD & $56.07\pm0.39$ & $56.12\pm0.32$ & $+0.05\pm0.69$ & $50.16\pm2.50$ & $0.00\pm0.00$ & 0/5/5 \\
FM-KD & $54.66\pm0.51$ & $54.29\pm0.59$ & $-0.37\pm0.84$ & $17.97\pm4.94$ & $45.00\pm3.61$ & 5/0/5 \\
\bottomrule
\end{tabular}
\end{table}

Relative to Appendix~\ref{app:training}, CIFAR-10 and MiniImageNet
teachers use learning rates $0.022593$ and $0.0408$, weight decays
$0.002049$ and $0.00181$, batch 128, warm-ups of four and one epochs,
watermark cadences of four and eight minibatches, and
$(\alpha,\gamma,\lambda_{\mathrm{aux}},\lambda_{\mathrm{orth}})
=(0.075346,8.049332,0.616596,1.564968)$ and $(0.4747,1.67,1.293,1.136)$.
Students use $(\tau,w_{\mathrm{CE}},\gamma_{\mathrm{FD}})=(4,0.5,10.515564)$
and $(3.683,0.19,30.78)$ in the objectives
$(1-w_{\mathrm{CE}})\mathcal L_{\mathrm{KD}}+w_{\mathrm{CE}}\mathcal L_{\mathrm{CE}}$
and
$(1-w_{\mathrm{CE}})\gamma_{\mathrm{FD}}\mathcal L_{\mathrm{FD}}+w_{\mathrm{CE}}\mathcal L_{\mathrm{CE}}$.

\subsection{Other Student Architectures}
\label{sec:crossarch-extension}

The first five teacher pairs are distilled into ResNet-50, VGG-16-BN,
MobileNetV3-Large and ViT-S/16 students
\citep{he2016resnet,simonyan2015vgg,howard2019mobilenetv3,dosovitskiy2021vit},
80 in total, with the student recipe of Appendix~\ref{app:training}. Convolutional
students start from random weights, whereas ViT-S/16 starts from ImageNet-21k
weights fine-tuned on ImageNet-1k \citep{5206848}, with $224\times224$ inputs and batch
32, so comparisons hold within each architecture. In every architecture,
the unmatched channel yields no detection, and the second-moment bound
certifies detection in 16 of 20 FM-KD students
(Table~\ref{tab:crossarch-all-channels}). Of the five ViT KL-KD students,
three reject through both tests, one only through the weighted test and
one through neither. Distilling the MiniImageNet teachers into the same
pretrained ViT-S/16 detects all ten joint students under dual access
(2.19\% logit BER after KL-KD, 13.44\% feature BER after FM-KD, with the
unmatched channels at 47.66\% and 44.53\%). Student accuracies of 36.5--39.4\%,
clean or marked, make this evidence of detection only.

\begin{table}[t]
\centering\scriptsize
\caption{Both readouts of the joint students of four architectures (CIFAR-100, five seeds). F/L BER: bit error rate in \% (mean$_{\pm\mathrm{SD}}$). F/L det.\ and Dual: detections under feature, logit or dual access. Cert.\ bits and Cert.\ det.: feature bits certified by the second-moment bound (of 64) and the detections they certify.}
\label{tab:crossarch-all-channels}
\begin{tabular}{llrrrrrrr}
\toprule
Student & Attack & F BER & F det. & L BER & L det. & Dual & Cert. bits & Cert. det. \\
\midrule
ResNet-50 & FM-KD & $3.44_{\pm 3.56}$ & 5/5 & $46.25_{\pm 3.42}$ & 0/5 & 5/5 & $53.60_{\pm 3.91}$ & 5/5 \\
ResNet-50 & KL-KD & $52.50_{\pm 8.31}$ & 0/5 & $5.62_{\pm 2.61}$ & 5/5 & 5/5 & $0.20_{\pm 0.45}$ & 0/5 \\
VGG-16-BN & FM-KD & $3.44_{\pm 3.39}$ & 5/5 & $50.00_{\pm 5.74}$ & 0/5 & 5/5 & $54.80_{\pm 3.19}$ & 5/5 \\
VGG-16-BN & KL-KD & $56.25_{\pm 5.95}$ & 0/5 & $0.62_{\pm 0.86}$ & 5/5 & 5/5 & $0.20_{\pm 0.45}$ & 0/5 \\
MobileNetV3-Large & FM-KD & $4.06_{\pm 3.60}$ & 5/5 & $46.88_{\pm 4.56}$ & 0/5 & 5/5 & $50.80_{\pm 6.14}$ & 4/5 \\
MobileNetV3-Large & KL-KD & $50.62_{\pm 6.50}$ & 0/5 & $22.81_{\pm 4.07}$ & 5/5 & 5/5 & $0.40_{\pm 0.55}$ & 0/5 \\
ViT-S/16 (pretrained) & FM-KD & $4.06_{\pm 3.60}$ & 5/5 & $46.88_{\pm 6.90}$ & 0/5 & 5/5 & $40.20_{\pm 6.87}$ & 2/5 \\
ViT-S/16 (pretrained) & KL-KD & $50.00_{\pm 7.25}$ & 0/5 & $27.81_{\pm 7.44}$ & 4/5 & 3/5 & $0.40_{\pm 0.55}$ & 0/5 \\
\bottomrule
\end{tabular}
\end{table}

\subsection{Removal Attacks on Extracted Students}
\label{sec:wrt-removal}
\label{sec:dehydra-removal}
\label{sec:sok-removal}

Five fixed attacks target the 40 CIFAR-100 students of clean and joint
teachers under both objectives, over all ten seeds and with identical code
for every seed. The attacker holds its student
and the training inputs, but not the key, the writers or the verifier.
Each recipe is fixed in advance, its last checkpoint is the endpoint, and
detection uses the four-test dual-access family at total level $10^{-3}$.
WRT weight shifting
\citep{lukas2022sok} replaces every convolution weight $W$ by
$W-1.5\,\overline W_{\mathrm{out}}-\epsilon_W$, with
$\overline W_{\mathrm{out}}$ the mean over output filters and Gaussian
$\epsilon_W$ at the tensor's standard deviation. It then fine-tunes with
fresh SGD for ten epochs on the student's own labels for 16,666 training
images, so it is not an unmodified WRT reproduction. FTAL trains
on true labels, and RTAL resets the classifier and then trains on the
student's soft outputs, both for five epochs at learning rate $0.01$. Fine-pruning zeroes the 96\% least active output activations of
the last convolutional stage and fine-tunes for ten epochs at learning rate
$10^{-3}$. Neural Dehydration \citep{10.1145/3658644.3690334} inverts each class
into 250 inputs by 800 Adam steps on
\[
\mathrm{CE}+0.01\,\mathcal{L}_{2}+0.03\,\mathcal{L}_{\mathrm{TV}}
+0.01\,\mathcal{L}_{\mathrm{BN}}.
\]
Here CE targets the inverted class, $\mathcal{L}_{2}$ and
$\mathcal{L}_{\mathrm{TV}}$ are the released image priors, and
$\mathcal{L}_{\mathrm{BN}}$ matches batch-norm statistics, a term the
release defines but omits from its loss. It then unlearns the recovered
inputs for 40 epochs with a uniform-target KL term (weight 15) beside
cross-entropy on 1,000 clean training images.

Every attack costs 11 to 19 accuracy points, in clean-teacher students as
much as in marked ones (Table~\ref{tab:removal-ten}). The feature channel
stays detected in every FM-KD student under all five attacks. Four attacks remove every KL-KD logit detection, while Neural
Dehydration leaves all ten logit-detected at 22.19\% logit BER, nine of
them under dual access. Fine-pruning flags one clean-teacher student, the
KL-KD student of seed 2718, under feature and dual access. It is the only
false flag among the 100 attacked clean students. These recipes show
vulnerability under fixed attacks, not a utility cost that any removal must
pay, and none retrains the comparison head that the feature verifier
reads.

\begin{table}[t]
\centering\scriptsize
\setlength{\tabcolsep}{3pt}
\caption{Removal attacks on the extracted CIFAR-100 students over all ten seeds (mean$_{\pm\mathrm{SD}}$ in \%). $\Delta$: paired accuracy change. F/L/D: feature, logit and dual-access detections at total level $10^{-3}$. Clean rows count false flags with the same-seed joint owner geometry.}
\label{tab:removal-ten}
\begin{tabular}{@{}lllrrrrrc@{}}
\toprule
Attack & Teacher & KD & Acc. before & Acc. after & $\Delta$ & F BER & L BER & F/L/D \\
\midrule
WRT weight shifting & joint & FM & $58.15_{\pm 0.36}$ & $43.24_{\pm 0.75}$ & $-14.91_{\pm 1.02}$ & $3.91_{\pm 2.88}$ & $48.59_{\pm 5.53}$ & 10/0/10 \\
 & joint & KL & $59.12_{\pm 0.36}$ & $43.27_{\pm 1.35}$ & $-15.85_{\pm 1.40}$ & $49.06_{\pm 7.45}$ & $47.66_{\pm 5.76}$ & 0/0/0 \\
 & clean & FM & $58.19_{\pm 0.27}$ & $43.52_{\pm 0.53}$ & $-14.67_{\pm 0.63}$ & $49.69_{\pm 6.15}$ & $49.22_{\pm 6.47}$ & 0/0/0 \\
 & clean & KL & $59.51_{\pm 0.38}$ & $44.02_{\pm 0.82}$ & $-15.49_{\pm 1.04}$ & $50.31_{\pm 7.02}$ & $50.62_{\pm 7.45}$ & 0/0/0 \\
\midrule
Neural Dehydration & joint & FM & $58.15_{\pm 0.36}$ & $40.07_{\pm 0.93}$ & $-18.08_{\pm 0.91}$ & $3.44_{\pm 3.52}$ & $48.75_{\pm 6.74}$ & 10/0/10 \\
 & joint & KL & $59.13_{\pm 0.36}$ & $41.31_{\pm 0.66}$ & $-17.82_{\pm 0.76}$ & $49.38_{\pm 6.68}$ & $22.19_{\pm 5.84}$ & 0/10/9 \\
 & clean & FM & $58.19_{\pm 0.27}$ & $40.28_{\pm 1.06}$ & $-17.91_{\pm 1.04}$ & $50.47_{\pm 7.11}$ & $48.91_{\pm 8.40}$ & 0/0/0 \\
 & clean & KL & $59.52_{\pm 0.38}$ & $41.76_{\pm 0.47}$ & $-17.76_{\pm 0.61}$ & $49.22_{\pm 6.56}$ & $49.06_{\pm 8.91}$ & 0/0/0 \\
\midrule
FTAL & joint & FM & $58.14_{\pm 0.36}$ & $40.76_{\pm 1.51}$ & $-17.38_{\pm 1.83}$ & $3.75_{\pm 2.68}$ & $47.66_{\pm 6.04}$ & 10/0/10 \\
 & joint & KL & $59.12_{\pm 0.36}$ & $40.23_{\pm 0.83}$ & $-18.89_{\pm 0.79}$ & $47.81_{\pm 6.39}$ & $45.16_{\pm 6.77}$ & 0/0/0 \\
 & clean & FM & $58.19_{\pm 0.26}$ & $41.17_{\pm 1.56}$ & $-17.01_{\pm 1.69}$ & $49.53_{\pm 6.21}$ & $47.81_{\pm 4.31}$ & 0/0/0 \\
 & clean & KL & $59.51_{\pm 0.38}$ & $41.04_{\pm 1.15}$ & $-18.47_{\pm 1.31}$ & $48.91_{\pm 7.87}$ & $49.06_{\pm 6.47}$ & 0/0/0 \\
\midrule
RTAL & joint & FM & $58.14_{\pm 0.36}$ & $44.06_{\pm 1.03}$ & $-14.08_{\pm 1.10}$ & $4.84_{\pm 3.25}$ & $49.53_{\pm 6.08}$ & 10/0/10 \\
 & joint & KL & $59.12_{\pm 0.36}$ & $43.78_{\pm 1.31}$ & $-15.35_{\pm 1.29}$ & $48.91_{\pm 6.67}$ & $46.09_{\pm 7.84}$ & 0/0/0 \\
 & clean & FM & $58.19_{\pm 0.26}$ & $43.60_{\pm 1.27}$ & $-14.58_{\pm 1.44}$ & $49.69_{\pm 5.55}$ & $50.94_{\pm 5.27}$ & 0/0/0 \\
 & clean & KL & $59.51_{\pm 0.38}$ & $44.16_{\pm 1.38}$ & $-15.35_{\pm 1.31}$ & $48.59_{\pm 6.89}$ & $50.78_{\pm 7.34}$ & 0/0/0 \\
\midrule
Fine-pruning & joint & FM & $58.14_{\pm 0.36}$ & $47.07_{\pm 0.77}$ & $-11.07_{\pm 0.80}$ & $10.16_{\pm 3.55}$ & $47.81_{\pm 3.62}$ & 10/0/10 \\
 & joint & KL & $59.12_{\pm 0.36}$ & $47.40_{\pm 0.95}$ & $-11.73_{\pm 1.02}$ & $47.50_{\pm 10.08}$ & $45.47_{\pm 6.61}$ & 0/0/0 \\
 & clean & FM & $58.19_{\pm 0.26}$ & $46.50_{\pm 1.10}$ & $-11.69_{\pm 1.10}$ & $47.81_{\pm 8.69}$ & $49.53_{\pm 7.62}$ & 0/0/0 \\
 & clean & KL & $59.51_{\pm 0.38}$ & $47.20_{\pm 1.73}$ & $-12.31_{\pm 1.97}$ & $48.28_{\pm 11.52}$ & $49.06_{\pm 7.77}$ & 1/0/1 \\
\bottomrule
\end{tabular}
\end{table}

\subsection{Negative Controls and Full-Key Calibration}
\label{sec:negative-controls}

Ten independently trained clean teachers and the students distilled from
them form the negative cohorts. Each model is scored with its own
comparison head and no payload-fitted alignment, threshold or head. With
one fresh, prespecified key per model, no cohort and no access scenario
yields a false detection (Table~\ref{tab:negative-controls}). Zero in ten
bounds a population false-positive rate only by 25.89\% (one-sided 95\%),
so these controls check the executed procedure, whose tests are exact or
conservative under the conditional null. A battery of 100 keys per model
flags a few queries, all from one key under feature or under logit access
and from those two keys under dual access, and every marked student lies
beyond every clean query (Figure~\ref{fig:null-separation}).
Its queries share models and keys, so its counts are not independent
trials. Only keys 28 and 67 ever flag, always through the vote test.
Under a fixed key, clean models give nearly the same logit scores (mean
pairwise cosine 0.96 to 0.98), so key 28 flags 18 of the 30 clean models.
The level thus holds on average over keys, not for each fixed key.

Full-key randomization (Appendix~\ref{app:full-key}) with 1,999 fresh,
precommitted keys reproduces every analytic decision on the 60 primary
models under all three access scenarios. Each joint teacher and matching
student reaches the rank floor $5\times10^{-4}$, and no clean model is
flagged (smallest $p_{\mathrm{key}}=0.009$).

\begin{table}[h]
\centering\small
\caption{False detections among independently trained clean models.
Entries give feature / logit / dual access, each at total level
$10^{-3}$. The 100-key battery reuses the same ten models per cohort, so
its 1,000 queries are correlated.}
\label{tab:negative-controls}
\begin{tabular}{lcc}
\toprule
Cohort & One key/model (of 10) & 100 keys/model (of 1,000) \\
\midrule
Clean teachers & 0 / 0 / 0 & 2 / 9 / 8 \\
Clean-teacher KL-KD students & 0 / 0 / 0 & 0 / 9 / 3 \\
Clean-teacher FM-KD students & 0 / 0 / 0 & 1 / 0 / 1 \\
\bottomrule
\end{tabular}
\end{table}

\section{Feature Transfer Beyond Classification}
\label{sec:downstream}
\label{sec:gnss-extension}
\label{sec:voc-extension}
\label{sec:isic-extension}

The feature writer needs only an exposed embedding, so it applies unchanged
to models without a classifier head. For each task, a marked and a clean
teacher train from the same initialization and budget. Each is distilled
into a student that trains its own 128-dimensional comparison head by
matching raw and normalized comparison features at unit weight, beside its
task loss where it has one. Both arms freeze the teacher and train the same
student parameters. A clean teacher exposes its untrained initial
comparison head, since no loss acts on it. Every model is audited by the feature-only test
at total level $10^{-3}$ on the first five seeds
(Table~\ref{tab:downstream}).

\begin{table}[ht]
\centering\small
\setlength{\tabcolsep}{3pt}
\caption{Feature-channel transfer beyond classification (five seeds, mean$\pm$SD in \%). Clean/Marked: utility of the clean and marked lineages, measured by per-image Dice, VOC2007 11-point mAP at IoU 0.5, or 5-way 5-shot episodic accuracy. BER: marked models' feature bit error rate. Det.: marked models detected by the feature-only test at total level $10^{-3}$. Flags: clean models detected.}
\label{tab:downstream}\label{tab:isic-paired}\label{tab:voc-paired}\label{tab:gnss-paired}
\begin{tabular}{llrrrcc}
\toprule
Task (utility) & Model & Clean & Marked & BER & Det. & Flags \\
\midrule
ISIC 2018 (Dice) & Teacher & $90.01\pm0.23$ & $89.37\pm0.21$ & $3.12\pm3.66$ & 5/5 & 0/5 \\
 & Student & $90.21\pm0.13$ & $90.17\pm0.11$ & $3.44\pm3.39$ & 5/5 & 0/5 \\
VOC07+12 (mAP) & Teacher & $76.86\pm0.22$ & $77.13\pm0.17$ & $4.06\pm3.24$ & 5/5 & 0/5 \\
 & Student & $20.90\pm0.53$ & $20.55\pm0.45$ & $7.81\pm6.54$ & 5/5 & 0/5 \\
GNSS/CRPA (acc.) & Teacher & $85.54\pm9.75$ & $92.50\pm0.64$ & $3.75\pm3.92$ & 5/5 & 0/5 \\
 & Student & $71.52\pm14.34$ & $87.46\pm4.31$ & $4.06\pm3.24$ & 5/5 & 0/5 \\
\bottomrule
\end{tabular}
\end{table}

\paragraph{Lesion segmentation.}
U-Nets \citep{10.1007/978-3-319-24574-4_28} with an ImageNet-initialized
EfficientNet-B4 encoder \citep{tan2019efficientnet} segment the ISIC 2018 Task~1 lesions
\citep{8363547,Tschandl2018}. The first 80\% of
the 2,594 images, sorted by name, train the models and the remaining 519
evaluate them, at $256\times256$. The writer reads globally pooled
deepest-stage encoder features ($K=64$, $r=16$, 16 carriers, 20 anchors,
hinge target $0.3$). Teachers and students train for 40 epochs with AdamW,
and utility is the mean per-image Dice score. The split is image-level
rather than patient-separated, and it is not the challenge test set.

\paragraph{Object detection.}
Swin-T \citep{liu2021swin} feature-pyramid \citep{8099589} Faster R-CNN
\citep{ren2016fasterrcnnrealtimeobject} detectors,
initialized from ImageNet-1k weights, train on the 16,551 PASCAL
VOC2007+2012 \citep{Everingham2010} trainval images.
Utility is the VOC2007 11-point mAP at intersection over union (IoU)
0.5 on the 4,952 test images. The writer reads the
768-dimensional pooled backbone features ($K=64$, $r=16$, hinge target
$0.3$). Teachers train for 15 epochs and randomly initialized students for
10 on $224\times224$ views, both with SGD. This short student recipe,
identical for both arms, reaches only about 20.5\% mAP against 77\% for
the teachers, so this task shows transfer to weak students, not protection
of a useful detector copy.

\paragraph{GNSS jammer detection and classification.}
Four-channel ResNet-18 models classify $4\times32\times32$ inputs from the
GNSS interference dataset of \citet{heublein_feigl_crpa}, recorded with a
four-element CRPA, into six classes for jammer detection and
classification (80,000 training and 10,000 evaluation inputs). Classification is 5-way 5-shot episodic with support-mean
prototypes \citep{3294996.3295163}, over 100 fixed episodes per seed
drawn from the training classes, so it measures episodic classification
rather than novel-class generalization. Teachers start from a
task-pretrained initialization, whose pretraining split is not recorded,
and train for 25 epochs with AdamW (64 carriers, 20 anchors). Students
train from random initialization by feature matching alone. Clean-teacher
students vary widely across seeds (SD 14.3 points), and the pretraining
split is unknown, so GNSS supports neither a utility nor a generalization
claim.

\section{Limitations and Broader Impact}
\label{app:limitations_outlook}
\label{app:broader_impact}

The boundaries below are measured rather than open-ended, and most of
them point to a concrete next step.

\paragraph{What the certificates cover.}
The certificates are sufficient conditions evaluated on saved outputs at
the audit inputs, and a small distillation loss on the adversary's own
inputs does not imply them (Section~\ref{sec:theory}). They explain an
observed recovery rather than guarantee survival over an attack class
fixed in advance. The feature
certificate holds for trained students at moderate payload sizes. It
certifies detection in every student up to 64 bits and in one of five at
128 bits, and beyond that detection rests on the calibrated tests
(Section~\ref{sec:experiments}). The logit bounds certify no bit of any
trained KL-KD student, although the stacked decoder is well conditioned
(Appendix~\ref{sec:certificates-key-only}). The KL
bridge also requires a floor on the served teacher's probabilities
(Corollary~\ref{cor:kl-floor}, Appendix~\ref{app:kl-floor-proof}).
Finally, Theorems~\ref{thm:bertest-cc} and~\ref{thm:capacity-cc} treat the
key expansion as independent randomness. Full-key randomization checks the
tests without this idealization (Appendix~\ref{app:full-key}), and each
realized design is certified by its measured rank and smallest singular
value, since the concentration bound need not be informative for large
payloads.

\paragraph{What the threat model covers.}
Each channel needs the output it reads. The logit channel needs full
logits, or full probabilities at a known temperature, together with
labeled audit inputs, and the feature channel needs the declared
comparison space (Appendix~\ref{app:notation}). The carrier inputs are
fixed training images and are not secret, so the owner must keep only the
key secret, but an input-selective wrapper can change the carrier
responses alone. Key probing, stripping the comparison space, overwriting the
mark, and hard-label or top-$k$ interfaces lie outside the evaluated
attacks (Section~\ref{sec:preliminaries}). Among the evaluated ones, an
orthogonal rotation of the served embedding hides the feature mark from a
key-only verifier, as Proposition~\ref{thm:rot}
(Appendix~\ref{app:readout-definitions}) predicts. An alignment fitted on
payload-bearing embeddings restores detection but also flags clean
teachers, so undoing a rotation needs reference material that is
independent of the payload (Appendix~\ref{sec:battery}). The feature
results therefore concern the declared coordinates and the responses on
the public carriers. The five fixed removal recipes erase the logit mark
while costing clean controls as much accuracy, and none retrains the head
that the feature verifier reads, so utility-preserving attacks on that head
remain untested (Appendix~\ref{sec:sok-removal}). Because the two writers share a
backbone, their errors need not be independent, and a targeted attack
could affect both channels.

\paragraph{What the experiments establish.}
All comparisons are fixed-recipe comparisons, because the recipe was
inherited from runs selected on evaluation accuracy, and ten seeds cannot
establish that two arms are equivalent (Appendix~\ref{app:training}).
Marking costs the CIFAR-100 teacher 1.29 accuracy points and about 2.3
times its training time. The primary students start from their teacher's
initialization as a matched control. Students of other architectures
share no initial weights with their teachers and still inherit the mark,
but a same-architecture control with an independent initialization would
isolate this factor more directly. At large payloads, detection alone is
weak evidence of a payload, because detection power grows with the
payload size, so signature-length claims also rest on measured rank and
bit error rates. They concern key-derived signatures, not capacity at a
fixed distortion (Appendix~\ref{sec:capacity-extension}). The one rank-deficient design, CIFAR-10, gives no logit
detection, although rank alone does not explain this, and one pretrained
ViT student escapes logit detection. Weak students limit some
conclusions. The MiniImageNet ViT results count as detection evidence
only (Appendix~\ref{sec:crossarch-extension}), the object-detection
results support only paired differences, and the GNSS results carry no
utility claim (Appendix~\ref{sec:downstream}). The baseline comparison
covers one distillation watermark, CosWM, without matching its distortion,
and evaluates no representation watermark such as SSL-WM. Ten independently trained
clean models per cohort give no false detection with one key per model,
but they bound a population false-positive rate only by 25.89\%
(one-sided 95\%). They check the executed procedure, whose tests are exact
or conservative under the conditional null, and the 100-key battery
reuses models and keys, so its counts are not independent trials
(Appendix~\ref{sec:negative-controls}). Its flags concentrate on a few
keys, so the level holds on average over keys, not for every fixed key. Fine-pruning also produced one
false flag among 100 attacked clean students, so null calibration should
cover attacked models as well.

\paragraph{Broader impact.}
TwinMark gives owners auditable evidence of model reuse that needs
neither the suspect's weights nor crafted trigger inputs, and its
certificates make explicit when a detection is guaranteed. A detector
rejection remains evidence relative to a declared null model, not a
determination of legal ownership. Repeated testing, keys chosen after
inspecting a suspect and correlated suspects all require their own
calibration. Fixed removal recipes can strip the logit mark, so owners
should not rely on a single channel. Audit data and model access must respect the privacy and
access requirements of each deployment.

\section{Licenses for Existing Assets}
\label{app:licenses_existing_assets}

We use only publicly available datasets, model architectures and open-source code assets, and we credit the original creators in the main paper and bibliography. Table~\ref{tab:asset_licenses} summarizes all licenses for existing assets. We do not repackage or redistribute restricted dataset assets. For datasets without an explicit upstream license, we list the official source, state that no license is specified and use the data only for research evaluation. ImageNet-derived datasets are treated under the ImageNet terms of access, which restrict use to non-commercial research and educational purposes.

All external assets are used only under their stated licenses or terms of use. We cite the original dataset and method papers, preserve license notices for open-source code dependencies, and exclude restricted datasets from the supplementary material and public source-code release. Our released code will contain only scripts, configuration files, and instructions for downloading public datasets from their original sources.

\begin{table}[!t]
\centering
\scriptsize
\setlength{\tabcolsep}{3pt}
\caption{Existing datasets and code/model assets used in the experiments, with source, license or terms of use, and access URL.}
\label{tab:asset_licenses}
\begin{tabular}{p{2.7cm} p{3.3cm} p{3.5cm} p{3.6cm}}
\toprule
\textbf{Asset} & \textbf{Source / repository} & \textbf{License or terms} & \textbf{URL} \\
\midrule
CIFAR-10~\citep{Krizhevsky2009LearningML} & Official Toronto CIFAR page & No explicit upstream license, used as a public research benchmark, not redistributed & \url{https://www.cs.toronto.edu/~kriz/cifar.html} \\
CIFAR-100~\citep{Krizhevsky2009LearningML} & Official Toronto CIFAR page & No explicit upstream license, used as a public research benchmark, not redistributed & \url{https://www.cs.toronto.edu/~kriz/cifar.html} \\
MiniImageNet~\citep{VinyalsBLKW16} & ImageNet-derived dataset and MiniImageNet generation tools & Underlying images follow ImageNet terms of access, and MiniImageNet tools are MIT licensed & \url{https://www.image-net.org/download.php}, \url{https://github.com/yaoyao-liu/mini-imagenet-tools} \\
PASCAL VOC 2007/2012 \citep{Everingham2010} & PASCAL VOC official dataset & Images obtained from Flickr, whose terms of use apply & \url{http://host.robots.ox.ac.uk/pascal/VOC/} \\
ISIC 2018~\citep{8363547} & ISIC Challenge archive & CC-BY-NC license for the aggregate ISIC 2018 training data, used for non-commercial research evaluation & \url{https://challenge.isic-archive.com/data/} \\
GNSS interference dataset~\citep{heublein_feigl_crpa} & Fraunhofer IIS GNSS GitLab, dataset for jammer detection and classification & CC BY-NC-SA 4.0 & \url{https://gitlab.cc-asp.fraunhofer.de/darcy_gnss/controlled_low_frequency} \\
\midrule
PyTorch & \texttt{pytorch/pytorch} & BSD-3-Clause & \url{https://github.com/pytorch/pytorch} \\
torchvision datasets and model implementations & \texttt{pytorch/vision} & BSD-3-Clause & \url{https://github.com/pytorch/vision} \\
ResNet-18/50~\citep{he2016resnet}, VGG-16-BN~\citep{simonyan2015vgg}, MobileNetV3~\citep{howard2019mobilenetv3} implementations & torchvision model zoo & BSD-3-Clause, via torchvision & \url{https://github.com/pytorch/vision} \\
ViT-S/16~\citep{dosovitskiy2021vit}, EfficientNet-B4, and related image backbones where instantiated through timm & \texttt{huggingface/ pytorch-image-models} (\texttt{timm}) & Apache-2.0 & \url{https://github.com/huggingface/pytorch-image-models} \\
Swin Transformer / Swin-T reference implementation & \texttt{microsoft/Swin- Transformer} & MIT & \url{https://github.com/microsoft/Swin-Transformer} \\
U-Net / segmentation components where instantiated through segmentation-models-pytorch & \texttt{qubvel-org/ segmentation\_models. pytorch} & MIT & \url{https://github.com/qubvel-org/segmentation_models.pytorch} \\
\bottomrule
\end{tabular}
\end{table}

\end{document}